\documentclass{article}
\usepackage{acl}

\usepackage[utf8]{inputenc}
\usepackage{times}
\usepackage{latexsym}

\usepackage[T1]{fontenc}
\usepackage[utf8]{inputenc}

\usepackage{microtype}
\usepackage{inconsolata}

\usepackage{booktabs}
\usepackage{amsfonts}
\usepackage{nicefrac}
\usepackage{xcolor}
\usepackage{graphicx}
\usepackage{amsmath}
\usepackage{multirow}
\usepackage{makecell}
\usepackage{array}
\usepackage{float}
\usepackage{changepage}
\usepackage{pgfplots}
\usepackage{caption}
\usepackage{subcaption}
\usepackage[most]{tcolorbox}
\usepackage{adjustbox}

\pgfplotsset{compat=newest}

\title{ZERA: Zero-init Instruction Evolving Refinement Agent\\
\large From Zero Instructions to Structured Prompts via Principle-based Optimization}
\author{Seungyoun Yi \\ 
    \And Minsoo Khang \vspace{0.5em} \\ Upstage AI Research \\
    \texttt{\{kyle, mkhang, sungrae.park\}@upstage.ai}
    \And Sungrae Park }

\begin{document}
\maketitle

\begin{abstract}
% Why prompt optimization is important or Short description of APO.
% Current limitation of the previous methods, 
% Short Introduction of ZERA
% Short Description of the results & findings
Automatic Prompt Optimization (APO) improves large language model (LLM) performance by refining prompts for specific tasks. However, prior APO methods typically focus only on user prompts, rely on unstructured feedback, and require large sample sizes and long iteration cycles—making them costly and brittle. We propose \textbf{ZERA} (Zero-init instruction Evolving Refinement Agent), a novel framework that jointly optimizes both system and user prompts through principled, low-overhead refinement. ZERA scores prompts using eight evaluation principles with automatically inferred weights, and revises prompts based on these structured critiques. This enables fast convergence to high-quality prompts using minimal examples and short iteration cycles. 
We evaluate ZERA across five LLMs and nine diverse datasets spanning reasoning, summarization, and code generation tasks. Experimental results demonstrate consistent improvements over strong baselines. Further ablation studies highlight the contribution of each component to more effective prompt construction.
Our implementation including all prompts is publicly available at \url{https://github.com/younatics/zera-agent}.

%Across five LLMs and nine diverse datasets—including reasoning, summarization, and code generation tasks—ZERA consistently improves performance over baselines. %Ablation studies show the independent benefits of structured evaluation, in-context examples, and prompt type coverage. 
%By unifying prompt critique and evolution under generalizable principles, ZERA offers a scalable, model-agnostic, and cost-efficient solution for robust prompt engineering.

% We introduce \textbf{ZERA} (Zero-prompt Evolving Reasoning Agent), a self-refining framework for automatic prompt optimization. Starting from minimal instructions (e.g., “You are a helpful assistant”), ZERA iteratively transforms prompts into task-sensitive formats through a multi-agent refinement loop.

% Unlike few-shot or handcrafted prompting, ZERA requires no exemplars or templates. It comprises two components: a \textit{Meta-Cognitive Agent} that proposes prompt updates based on interaction history, and an \textit{Evaluation Agent} that assigns task-adaptive scores. This structure induces semantically and structurally aligned prompts without external supervision.

% ZERA improves performance across six LLMs and eight benchmarks, especially on structured reasoning and code generation tasks. Ablation studies highlight the independent contributions of reasoning steps and in-context examples. Our results suggest that ZERA offers a scalable, model-agnostic path to automatic prompt engineering.
\end{abstract}
\section{Introduction}

The effectiveness of LLMs significantly depends on the quality of prompts used to guide their behavior. Crafting effective prompts is essential not only for general LLM application but also crucial when integrating LLMs into larger agent-based systems. However, developing these prompts typically relies on handcrafted templates, domain intuition, or extensive trial-and-error processes, which pose considerable challenges in scalability and transferability~\cite{brown2020language, perez2021true, zhao2021calibrate}. Moreover, optimal prompts are often model-specific, necessitating careful tuning of prompts to the particular LLM being employed.

To address these challenges, automatic prompt optimization (APO) methods have recently been proposed. The core objective of these approaches is to systematically derive prompts that yield desired outputs for given inputs in a specific task. This typically involves an iterative process where an LLM evaluates the effectiveness of a prompt, identifies shortcomings, and incrementally updates the prompt to enhance performance~\cite{wang2024promptagent,yang2024large,He_Liu_Xu_Shivade_Zhang_Srinivasan_Kirchhoff_2025}. However, these methods predominantly rely on task-specific metric scores and feedback derived solely from the provided examples, making them prone to overfitting and limiting their robustness in generalization.

% These methods significantly contribute to achieving task-specific goals tailored explicitly to the LLM used. Nevertheless, these methods predominantly optimize prompts solely based on provided task examples, there is a potential risk of overfitting to those examples. Consequently, these optimized prompts may lack robustness, limiting their practical applicability in real-world scenarios. 
% % [추가 paragraph] ZERA에 대한 소개가 들어가기 전에 너무 바로 도입되는 느낌입니다. OPRO, CriPo등의 방법이 어떤 목표와 어떤 구조를 가지고 위의 정의한 문제에 기여해 왔다는 내용이 먼저 있고, 하지만 여전히 개선될 수 있는 룸이 있다는 내용이 있으면 좋겠네요. - 민수님

To mitigate this limitation, we propose ZERA (Zero-init instruction Evolving Refinement Agent), a novel APO approach designed to improve the generality and robustness of optimized prompts. Instead of relying solely on task-specific feedback or metric scores derived from a small set of examples, ZERA employs eight evaluation principles for prompt optimization: Completeness, Conciseness, Correctness, Expression Style, Faithfulness, Meaning Accuracy, Reasoning Quality, and Structural Alignment. These principles serve as high-level evaluation criteria that guide feedback generation and prompt refinement, enabling the system to generalize beyond individual examples and avoid overfitting.

Specifically, ZERA consists of two iterative stages: Principle-based Critique Generation (PCG) and Meta-cognitive Prompt Refinement (MPR). PCG utilizes task-specific sample data to (1) evaluate the relative importance of each principle for a given task and (2) measure performance against each principle, generating output analysis and actionable feedback. MPR integrates this feedback to iteratively refine task-related meta-information, including task descriptions and the targeted optimization objectives—system and user prompt.

The iterative interaction between these two stages based on the meta principles results in the development of highly optimized system and user prompts. Notably, ZERA can generate effective prompts even when provided with only a few task samples and no handcrafted prompts or task descriptions. Furthermore, because task evaluation and definition are driven by general principles, the optimized prompts exhibit resistance to overfitting. Additionally, the influence of these general principles promotes rapid convergence during the prompt optimization steps, demonstrating ZERA’s practicality and effectiveness. 

We validated our proposed method across nine benchmark tasks—MMLU, MMLU-Pro, GSM8K, MBPP, HumanEval, BBH, HellaSwag, CNN/DM, and Samsum—optimizing prompts for models such as GPT-3.5, GPT-4o, LLaMA-3.1-70B-Instruct (LLaMA-3.1), Qwen-2.5-70B-Instruct (Qwen-2.5), and Mistral-7B-Instruct-v0.3 (Mistral-7B). In most cases, ZERA-derived prompts outperformed predefined prompts provided for each task. Additionally, we compared ZERA with recent APO methodologies, including PromptAgent~\cite{wang2024promptagent}, OPRO~\cite{yang2024large}, and CriSPO~\cite{He_Liu_Xu_Shivade_Zhang_Srinivasan_Kirchhoff_2025}, and observed that ZERA delivered superior performance. Furthermore, we conducted an ablation study to analyze the distinct characteristics and effectiveness of individual components within our proposed approach.

\section{Related Work}

% Task desciption 관련 내용
% 이 부분은 keep
A wide range of methods have been proposed in the field of APO, broadly categorized by whether they require training or gradient updates ~\cite{chen2024mapo, Zhang_Sang_2025, jafari-etal-2024-morl, Chen_Wang_Jiang_Nakashima_2025, srivastava2025revisiting}, or operate in a training-free manner ~\cite{He_Liu_Xu_Shivade_Zhang_Srinivasan_Kirchhoff_2025, xiang2025self, peng2025dlpo, wang2024promptagent, pryzant-etal-2023-automatic}. Training-based approaches offer the advantage of task-specific optimization through reinforcement learning or supervised tuning, often leading to higher performance on narrowly defined tasks. In contrast, training-free methods are more readily adaptable to new tasks, as they eliminate the computational and data requirements associated with model training.

% 이 부분은 기존 optimization방법들중, single-scalar score기반 feedback을 활용해서 prompt를 optimizing하는 work들에서
% natural language text 로 흘러가는 work들 설명
Among training-free methods, one of the earlier notable works is APE ~\cite{zhou2023large} which iteratively generates prompt variants and selects the best prompt based on task-specific metric scores. While effective, the use of scalar feedback offers limited guidance for understanding why a prompt is better or how to improve it further. To address this, subsequent works such as ~\cite{pryzant-etal-2023-automatic, peng2025dlpo, wang2024promptagent} enhance the optimization process by incorporating natural language feedback derived from error examples. These textual signals offer more descriptive and interpretable suggestions, guiding the LLM to generate improved prompts through enriched context.

% 이 부분은, natural language feedback을 넘어서 다양한 feature들을 추가하면서 더 고도화하는 work들 설명
Building on this trajectory, more recent approaches such as OPRO~\cite{yang2024large} and CriSPO~\cite{He_Liu_Xu_Shivade_Zhang_Srinivasan_Kirchhoff_2025} further enhance prompt optimization by incorporating additional signals beyond natural language feedback. OPRO stabilizes the optimization process by leveraging historical prompt traces, while CriSPO introduces a multi-aspect critique–suggestion agent that provides aspect-specific feedback. These innovations enable more targeted and robust improvements in prompt quality across iterations.

% 여기서 ZERA에서 사용된 8개의 principle들을 활용해서 user-prompt만 아닌 system & task description prompt들을 모두 optimize해서 차별화 하는지 설명
While ZERA incorporates common strategies from prior work, such as natural language feedback and historical prompt traces, it distinguishes itself by grounding the optimization process in eight generalizable principles. These principles guide structured feedback and drive the joint optimization of the user prompt, system prompt, and task description—components that are typically fixed or ignored in earlier approaches. To the best of our knowledge, ZERA is the first to unify the optimization of all three prompt types (system prompt, user prompt and task description) within a principle-driven framework.

% Building on this trajectory, more recent work such as OPRO ~\cite{yang2024large} introduces the use of historical prompt traces, enabling more stable optimization by preserving contextual continuity across iterations. Our work, ZERA, extends this line of training-free prompt optimization by enhancing the quality and structure of feedback through a set of well-defined principles (e.g., correctness, reasoning, format-following), providing more consistent and effective guidance for prompt refinement.

% Relatedly, CriSPO~\cite{He_Liu_Xu_Shivade_Zhang_Srinivasan_Kirchhoff_2025} explores a similar direction by introducing multi-aspect critique, where a critique LLM identifies multiple dimensions for improvement and provides suggestions for each. While aligned in its goal of enriching feedback for prompt optimization, ZERA distinguishes itself by employing a well-defined set of principles to guide feedback generation, enabling a more controlled and consistent optimization process.

\section{Methodology}
\label{sec:method}

ZERA approaches prompt optimization through an iterative, training-free framework comprising two key stages: evaluation and refinement. This section introduces the APO formulation and details the core components of ZERA: principle-based evaluation and meta-cognitive refinement modules, which work together to iteratively improve prompts from generic initial prompts.

\begin{figure*}[t!]
\centering
\includegraphics[width=0.95\linewidth]{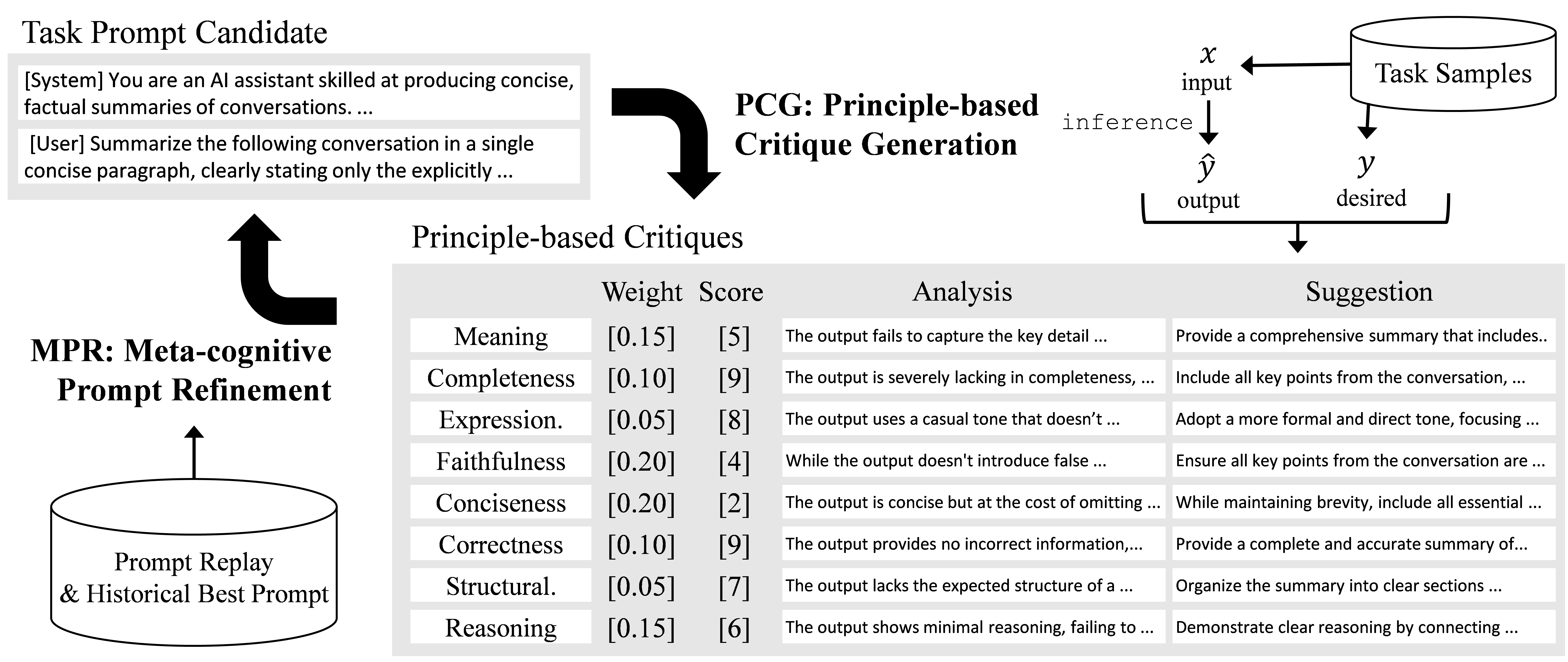}
\caption{Overview of the ZERA system. Given task samples and their corresponding output results, PCG produces the critique comprising of: importance weight, evaluation score, analysis result and suggestion across eight principles. MPR refines the task prompt by integrating prior prompt information with the critiques observed in the current task examples, along with historical feedback from previous iterations (Eq.~\ref{eq:history}).}
\label{fig:zera-arch}
\vspace{-5pt}
\end{figure*}

\subsection{Problem Formulation}

We begin by formalizing the prompt optimization objective and outlining its core challenges. We define a task  $\mathcal{D}$ as a set of paired examples $(x, y)$, where $x$ is the raw input and $y$ is the desired output. In prompt-based learning, LLMs do not consume $x$ directly; rather, it is embedded into a textual prompt $p_{\text{task}}$ that conditions the model's output. We denote the output of the LLM as $\hat{y} = \text{LLM}(x |\; p_{\text{task}}$).

The objective of APO is to find an optimal prompt function $p_{\text{task}}$ that minimizes the expected distance between model output and the ground-truth label:
\begin{equation}
    J(p_{\text{task}}) = \mathbb{E}_{(x, y) \sim \mathcal{D}} \left[ \text{dist}(\text{LLM}(x |\; p_{\text{task}}), y) \right].
\end{equation}
% A LLM task can be defined as as a set of tuples consisting of input ($x$) and its desired output ($y$) text when $(x, y)\in \text{TASK}$. Let denote $\hat{y}=\text{LLM}(x|p_{\text{task}})$ as the output of task LLM utilizing a prompt ($p_{\text{task}}$), then the objective function of APO can be defined as
% \begin{equation}
%     J(p) = \mathbb{E}_{(x,y)\sim \text{TASK}}\left [ \text{dist}(\text{LLM}_{\text{task}}(x|p_{\text{task}}), y)\right] 
% \end{equation}
Here, $\text{dist}(\hat{y}, y)$ denotes a distance metric quantifying the discrepancy between the LLM-generated output $\hat{y}$ and the ground-truth target $y$. The goal of APO is to identify an optimal prompt function $p^{*}_{\text{task}}$ that minimizes this objective. However, there are three key challenges in APO: (1) the LLM is typically accessed as a black box, offering no gradient or parameter-level information; (2) optimizing $p_{\text{task}}$ is non-trivial, as traditional gradient-based methods are inapplicable without model retraining; and (3) the available data for $\mathcal{D}$ is often limited, posing a challenge for generalization.

\subsection{Design Rationale}
To address these challenges, ZERA adopts the following design choices. As the first two challenges make it infeasible to directly optimize prompts using task-specific objectives, prior work has explored training-free alternatives for prompt optimization. Recent work~\cite{wang2024promptagent, He_Liu_Xu_Shivade_Zhang_Srinivasan_Kirchhoff_2025}, for example, introduces natural language-based feedback and optimization frameworks, where LLM-generated qualitative feedback is used to guide prompt refinement. Following this line of research, ZERA leverages natural language feedback as the central supervision signal for iteratively improving prompts.

The third challenge poses a significant risk to generalization. When prompt optimization relies heavily on LLM-generated evaluations and feedback, there is a heightened risk that prompts may overfit to a small set of biased or unrepresentative examples. To address this issue, our method introduces meta-level principles that serve as high-level guides for evaluation and feedback. By grounding prompt updates in these general reasoning frameworks, rather than just task-specific signals, our approach promotes broader applicability and reduces the risk of overfitting.

Given these constraints, prompt optimization is best approached in a heuristic, training-free framework composed of two iterative stages: evaluation and refinement. These two stages can be described as follows:
\begin{align}
    &\hat{\mathbf{y}}^{(t)} \leftarrow \text{LLM}_{\text{task}} \left( p_{\text{task}}^{(t)}(\mathbf{x}^{(t)}) \right) \\
    &\mathbf{c}^{(t)} \leftarrow \mathcal{A}_{\text{eval}} \left( \mathbf{x}^{(t)}, \hat{\mathbf{y}}^{(t)}, \mathbf{y}^{(t)} \right) \\
    &p_{\text{task}}^{(t+1)} \leftarrow \mathcal{A}_{\text{refine}} \left( \mathbf{x}^{(t)}, \hat{\mathbf{y}}^{(t)}, \mathbf{y}^{(t)}, \mathbf{c}^{(t)}, p_{\text{task}}^{(t)} \right)
\end{align}
Here, $\mathbf{x}^{(t)}$ and $\mathbf{y}^{(t)}$ denote the input and reference output sets at iteration $t$, and $\hat{\mathbf{y}}^{(t)}$ is the set of outputs generated by the task LLM using the current prompt $p_{\text{task}}^{(t)}$. The evaluation agent $\mathcal{A}_{\text{eval}}$ produces critique tuples $\mathbf{c}^{(t)}$ containing natural language suggestions and scores grounded in the meta-level principles, to assess the quality of the generated outputs. These critiques, along with the original inputs, outputs, and prompt, are then passed to the prompt modification agent $\mathcal{A}_{\text{refine}}$, which generates an updated prompt $p_{\text{task}}^{(t+1)}$.

As this formulation highlights, the effectiveness of the overall optimization process depends critically on the design of both $\mathcal{A}_{\text{eval}}$ and $\mathcal{A}_{\text{refine}}$, which determine how feedback is generated and how prompts are refined.

% Consequently, prompt optimization under constrained settings is best approached as a heuristic optimization framework composed of two recurring stages: evaluation and update, as shown as following.
% \begin{align}\
%     & \hat{\textbf{y}}^{(t)} \xleftarrow{} \text{LLM}_{\text{task}} \left ( \textbf{x}^{(t)} \middle| \; p^{(t)}_{\text{task}} \right ) \\
%     &\textbf{c}^{(t)} \xleftarrow{} \text{A}_{\text{eval}}\left ( \textbf{x}^{(t)}, \hat{\textbf{y}}^{(t)}, \textbf{y}^{(t)} \right ) \\
%     &p^{(t+1)}_{\text{task}} \xleftarrow{} \text{A}_{\text{prom}}\left ( \textbf{x}^{(t)}, \hat{\textbf{y}}^{(t)}, \textbf{y}^{(t)}, \textbf{c}^{(t)}, p^{(t)}_{\text{task}} \right ) 
% \end{align}
% where $\textbf{x}^{(t)}$ and $\textbf{y}^{(t)}$ denote sets of task inputs and pre-defined desired outputs, respectively. $\hat{\textbf{y}}^{(t)}$ indicates a set of generated outputs by $\text{LLM}_{\text{task}}$ with the current prompt, $p^{(t)}_{\text{task}}$. In the second line, $\text{A}_{\text{eval}}$ represent an agent system generating critiques, $\textbf{c}^{(t)}$, of generated task outputs. In the third line, $\text{A}_{\text{prom}}$ indicates an agent system updating the task prompt. As shown in the formulation, the effectiveness of prompt optimization critically depends on how $\text{A}_{\text{eval}}$ and $\text{A}_{\text{prom}}$ are designed. 

\subsection{System Overview and Principles}
As illustrated in Figure~\ref{fig:zera-arch}, ZERA follows a two-stage iterative process of evaluation and refinement. While structurally similar to conventional APO frameworks, it uniquely integrates principle-based evaluations to assess the current prompt and guide its refinement. The motivation for this design is to incorporate pre-defined meta-level information—namely, a set of general principles—to reduce the risk of bias that may arise when optimizing prompts from a limited number of task examples. 

%While following an iterative two-stage prompt optimization process, ZERA further enhances efficiency and robustness in both critique and prompt generation by incorporating generalizable principles and continuously updating task descriptions. 
% The retionale behind these additions it to inject meta-level information--derived from just a few task examples--to support the generation of stable and generaliable prompts. By leveraging these meta information, two agents share task descriptions, ensuring a consistent understanding of the task in the perspective of the principles. This, in turn, helps maintain a unified tone and objective across both the evaluation and prompt refinement stages, leading to more coherent and aligned optimization outcomes. 

Based on our analysis across diverse benchmark tasks, we identified eight generalizable principles that consistently guided effective prompt evaluation and refinement. These principles were inductively derived from recurring evaluation criteria observed in summarization, translation, and reasoning tasks, and are grounded in cognitive science (e.g., Bloom’s taxonomy), linguistic pragmatics (e.g., Gricean maxims), and NLP evaluation rubrics (e.g., factuality, fluency, coherence). Designed to balance coverage, generality, and interpretability, they enable ZERA to operate across diverse tasks without relying on handcrafted instructions or dataset-specific scoring rubrics. Summarized in Table~\ref{tab:principles}, they form the foundation for assessing and improving prompts, and are systematically applied by both PCG and MPR to ensure coherence and consistency throughout the optimization process.

% Figure~\ref{fig:zera-arch} shows an overview of ZERA system architecture. As can be seen, ZERA consists of two agents: Principle-based Critique Agent and Meta-cognitive Prompt Generation Agent. Two agents refer the same principles and task description at each iteration. 

% Similar to OPRO and CriSPO, ZERA adopts two-agent iterative process but each agent refers pre-defined principles and identified task descriptions. 
% In each iteration, the Principle-based Critic Agent evaluates the output of sampled examples in the perspective of the principles, while the Meta-cognitive Prompt Generation Agent updates the task description and prompt based on the feedback.

% ZERA is a self-contained, self-bootstrapping prompt optimization system that begins from null or underspecified prompts and progressively refines them through a meta-learning-inspired, multi-agent feedback loop (Figure~\ref{fig:zera-arch}). 
%Unlike prior approaches, it does not rely on handcrafted templates, few-shot exemplars, or task-specific heuristics.

% \begin{figure}[t!]
% \centering
% % \includegraphics[width=0.95\columnwidth]{sections/images/zera_architecture_diagram.png}
% \includegraphics[width=0.95\columnwidth]{sections/images/ZERA_Concept.png}
% \caption{Overview of the ZERA architecture. The Evaluation Agent scores responses across eight principles. The Meta-Cognitive Agent refines prompts based on prior outputs and evaluation feedback.}
% \label{fig:zera-arch}
% \end{figure}

\begin{table*}[ht!]
    \caption{Short description of eight principles. The detailed criteria  be found in Appendix A1.}
    \label{tab:principles}
    \small
    \centering
    \vspace{-2pt}
    \begin{tabular}{l|l}
        \toprule
         Principle &	Description \\
         \midrule
        Meaning Accuracy &	Preserves intended meaning and logical consistency with the expected answer (output fidelity).  \\
        Completeness &	Includes all key ideas or steps; no critical elements are missing.  \\ 
        Expression Style &	Matches tone, format, and stylistic elements of the expected answer.  \\
        Faithfulness &	Avoids hallucination; stays true to given input and context.  \\
        Conciseness &	Maintains brevity; avoids unnecessary or repetitive content.  \\
        Correctness &	Final answer is factually/logically correct and meets formatting constraints.  \\
        Structural Alignment &	Matches the structure, formatting, and layout of the expected answer.  \\
        Reasoning Quality &	Provides logically sound \& well-structured reasoning process aligned with task goals.   \\ \bottomrule
    \end{tabular}
    \vspace{-5pt}
\end{table*}

% \subsection{Multi-Agent Refinement Loop}

% At the heart of ZERA is an iterative feedback loop between two cooperating agents: the Evaluation Agent and the Meta-Cognitive Agent. We formulate this process as a bi-level optimization problem, where the Meta-Cognitive Agent learns to generate increasingly effective prompts by interpreting structured evaluation signals and analyzing historical generation patterns.

% \paragraph{Optimization Objective.}
% ZERA aims to maximize the alignment between the language model’s outputs and task-specific evaluation objectives, relative to gold-standard answers. Formally, we define the learning objective in Equation~\ref{eq:zera-objective}.

% \begin{equation}
% \label{eq:zera-objective}
% \max_{\theta, \phi} \sum_{t=1}^{T} 
% \sum_{k=1}^{8} \alpha_k^{(\phi)}(T_t)\, 
% V_k^{(\phi)}(\text{LLM}(f_\theta(H_{t-1}), x_t), y_t)
% \end{equation}

% The above objective quantifies how well the model's outputs align with task-specific criteria, as judged by a learned weighting mechanism. ZERA begins from structurally naive prompts (e.g., \texttt{"You are a helpful assistant"}) and progressively evolves them into structurally precise forms—prompts that elicit scorable, criterion-aligned outputs. This transformation is essential for enabling reliable feedback and meaningful refinement even when initial prompts are underspecified or unscorable.

\subsection{Principle-based Critique Generation (PCG)}
%\paragraph{Evaluation Agent and Score Decomposition.}
%Given an input $x^{(t)}$, a prompt $p^{(t)}_{\text{task}}$, and a reference answer $y^{(t)}$, the model produces an output $\hat{y}^{(t)} = \text{LLM}(P_t, x_t)$. 

Given the task inputs $\mathbf{x}^{(t)}$ and the corresponding LLM outputs $\hat{\mathbf{y}}^{(t)}$ generated using the current prompt $p^{(t)}_{\text{task}}$, the evaluation agent $\mathcal{A}_\text{eval}$ produces a detailed assessment and feedback for each sample. Our proposed PCG structures this process around a set of eight general principles, producing four key outputs. First, it analyzes the task description to estimate the relative importance of each principle, assigning a real-valued weight in the range [0-1] to reflect its priority. Second, it evaluates the generated outputs against each principle, producing a score [1-10] per principle to reflect output quality. Third, it conducts an error analysis to determine which aspects of the outputs were well-handled or problematic based on the eight principles. Lastly, it outputs targeted suggestions for improvement aligned with each principle.

% Given task inputs $\textbf{x}^{(t)}$ and these corresponding LLM outputs $\hat{\textbf{y}}^{(t)}$ with a prompt $p^{(t)}_{\text{task}}$, the evaluation agent, $\text{A}_\text{eval}$, produces a detailed assessment and feedback for each. The Principle-based Critique Agent structues this process using a principle-driven framework, yielding four key outputs. First, it analyzes the current task description to estimate the relative importance of each of the eight general principles, assigning a scalar weight [1-10] to each. Second, using each principle as evaluation criteria, it evaluates the outputs generated from the task samples and produces a scalar score [1-10]. Third, it analyzes the generated results to identify which aspects were handled well or poorly, and provides guidance on how to improve, with feedback organized by each individual principle. 
%For each task sample, it produces an eight-dimensional evaluation scores and constructs targeted feedback for further improvement. 

% c_n <- n번째 task sample에 대한 feedback
% 이 피드백안에는
% (1) 해당 샘플로 보았을 때 principle의 중요도 - alpha
% (2) princile 별 평가 점수 - s
% (3) printicle 별 분석 결과 (\hat{y}, y - 사이의 차이가 무엇인지 도출) - a
% (4) principle 별 action item - 어떻게 y_hat 이 바뀌면 좋겠다. - f

For clarity and formalization, we define the critique tuple for the $n$-th task sample as $c_n = (\alpha_n, s_n, a_n, f_n)$. Here, $\alpha_n$ is an eight-dimensional vector representing the estimated importance weights of the eight principles, and $s_n$ denotes the corresponding evaluation scores assigned to the generated output. The component $a_n$ captures the qualitative analysis of the output with respect to each principle, while $f_n$ provides principle-specific suggestions for improvement.

% For clarity and formalization, we denote the critique tuple for the $n$-th task sample result as $c_n = (\alpha_n, s_n, a_n, f_n)$. In this tuple, $\alpha_n$ represents an eight-dimensional vector of principle importance values, and $s_n$ denotes the corresponding evaluation scores. The component $a_n$ provides an analysis of the result with respect to each of the eight principles, while $f_n$ delivers improvement guidelines from the perspective of those principles. 
The critique tuple at time $t$ can be identified through the following:
% For formal description, we denote $\alpha^{(t)}$ be a eight-dimentional vector indicating the principle importance and $c^{(t)}_{n,k}$ be a critique text including a score for $n$-th task sample in the perspective of $k$-th principle. The two key outputs can be identified through the following process:
% \begin{align}
%     & \alpha^{(t)}_{n} \xleftarrow{} \text{LLM}_\text{eval} \left ( \text{T}^{(t)}_\text{task} \middle|\; p_{\alpha} \right ), \\
%     & c^{(t)}_{n,k} \xleftarrow{} \text{LLM}_\text{eval} \left ( \hat{y}^{(t)}_{n}, y^{(t)}_{n} \middle|\; p_{\text{critic}}, p_{\text{prin.}, k} \right ).
% \end{align}
\begin{equation}
    c^{(t)}_{n} \xleftarrow{} \text{LLM}_\text{eval} \left (  \text{T}^{(t)}_\text{task}, \hat{y}^{(t)}_{n}, y^{(t)}_{n}, x^{(t)}_{n} \middle|\; p_{\text{eval}} \right ).
\end{equation}
Here, $\text{T}^{(t)}_\text{task}$ denotes the task description at the $t$-th iteration, and $p_{\text{eval}}$ corresponds to the critique generation prompt including predefined principle definitions. Note that the task prompt, $p_{\text{task}}$, is not directly utilized in this stage; rather, the outputs generated from it on task samples are evaluated through the lens of the predefined principles. 

\subsection{Meta-cognitive Prompt Refinement (MPR)}
%\paragraph{Historical Context and Memory Encoding.}
In the prompt refinement stage, the core objective is to update the task prompt using the structured feedback produced during evaluation. Central to this process is our Meta-cognitive Prompt Refinement Agent, which leverages multi-dimensional, principle-based evaluations to guide refinement. By identifying which principles are most critical to the task—based on scalar importance scores—the agent redefines the task description and adjusts the prompt accordingly. This principled approach ensures that the updated prompts align with high-level quality dimensions such as reasoning, accuracy, and structure, making it the primary driver of generalizable and task-aligned prompt improvement.

To further stabilize and enhance the refinement process, the agent also incorporates historical information from past iterations. It considers (1) recent prompts and their evaluation results, (2) the best-performing prompt to date and its scores, and (3) exemplar task samples—specifically, the three with the highest scores and two with the lowest. These historical references provide meta-level context, helping the agent maintain consistent progress, avoid local optima, and balance prompt quality across a range of task instances. While secondary to principle-based feedback, incorporating the historical information trajectory enhances optimization stability by enabling the model to avoid prior errors and reinforce effective strategies \cite{He_Liu_Xu_Shivade_Zhang_Srinivasan_Kirchhoff_2025}.

For formal description, let $\text{F}^{(t)}_{\text{task}}$ be a tuple of $(p^{(t)}_\text{task}, \textbf{c}^{(t)})$, indicating the task prompt feedback at the $t$-th iteration. For the task sample, we denote $\text{F}^{(t)}_{\text{sample}}$ as a tuple of $(\textbf{x}^{(t)}, \hat{\textbf{y}}^{(t)}, \textbf{y}^{(t)}, \textbf{c}^{(t)})$, corresponds to the task sample feedback at the $t$-th iteration. Using this definitions, the task prompt and description refinement can be described as the follow:
\begin{align}
\label{eq:history}
    p^{(t+1)}_{\text{task}},  & \text{T}^{(t+1)}_\text{task} \xleftarrow{}  \text{LLM}_{\text{refine}} {\LARGE (} \text{T}^{(t)}_\text{task}, \text{F}^{(t)}_{\text{task}}, 
    \text{F}^{(t-1)}_{\text{task}},
    \text{F}^{(t-2)}_{\text{task}}, \nonumber \\
    & \;\;\;\;\;\;\;\;\;\; \text{F}^{(t), *}_{\text{task}}, \text{F}^{(t), \text{top-3}}_{\text{sample}}, \text{F}^{(t), \text{bottom-2}}_{\text{sample}}  {\LARGE |}\; p_{\text{refine}} {\LARGE )},
\end{align}
where $\text{F}^{(t), *}_{\text{task}}$ represents the tuple showing the best feedback score among all previous iterations. $\text{F}^{(t), \text{top-3}}_{\text{sample}}$ and $ \text{F}^{(t), \text{bottom-2}}_{\text{sample}}$ indicate the top three and the bottom two of task sample feedback along with the evaluated scores at the iteration $t$. By combining the current task and sample feedback and the historical records, $\text{LLM}_{\text{refine}}$ refines the task prompt and description. 
%The prompt, $p_{\text{prom}}$, used in this stage can be found in Appendix XX.

Since our evaluation is based on multiple principles, it naturally produces multi-dimensional scores for each output. To identify the best and worst prompt cases in the historical data, we compute a unified score that integrates these dimensions. This aggregation relies on the principle importance weights generated during the evaluation stage, allowing the system to weigh each criterion according to its relevance to the task. In other words, for each sample, the unified score, $u^{(t)}_{n}$ is calculated as follows:
\begin{equation}
\label{eq:unified-score}
    u^{(t)}_{n} = \sum_{k} \alpha^{(t)}_{n,k} s^{(t)}_{n, k}, 
\end{equation}
where $\alpha^{(t)}_{n,k}$ represents the principle importance ratio and $s^{(t)}_{n, k}$ indicates the evaluation score of $\hat{y}^{(t)}_n$ in the view of the $k$-th principle. These scores can be identified from the critique tuple  $c^{(t)}_{n, k}$. The weighting vector $\alpha$ is adaptively determined based on the characteristics of each task and sample, allowing the system to assess the relative importance of different principles. As a result, the multi-dimensional evaluation scores are aggregated in a way that reflects what matters most for the specific task. For instance, in tasks where reasoning is not a critical factor, the weight assigned to the reasoning principle will be low. Consequently, scores related to reasoning will have minimal influence in identifying strong or weak task cases or in guiding prompt refinement.

\begin{table*}[t]
% Our Setting - LLM{Eval, Refine} 하는 것이 다르고 LLM{task}는 같음.
\centering
\caption{Performance across BBH subcategories. All results are re-evaluated under a consistent setting: GPT-3.5-turbo is used as the base model for response generation, and GPT-4o is used for prompt refinement where applicable (e.g., PromptAgent and ZERA). *Object Counting score for PromptAgent is taken from the original paper.}
\vspace{-2pt}
\label{tab:bbh-subtasks}
\renewcommand{\arraystretch}{1.2}
\small
\setlength{\tabcolsep}{3.8pt}
\begin{tabular}{lcccccc|c}
\toprule
Method & Penguins & Geometry & Epistemic & Object Count & Temporal & Causal Judge & Avg. \\
\midrule
Human (0 shot)        & 0.595 & 0.227 & 0.452 & 0.612 & 0.720 & 0.470 & 0.513 \\
CoT (0 shot)          & 0.747 & 0.320 & 0.532 & 0.542 & 0.734 & 0.610 & 0.581 \\
% GPT Agent         & 0.696 & 0.445 & 0.406 & 0.502 & 0.794 & 0.520 & 0.561 \\
% APE~\cite{He_Liu_Xu_Shivade_Zhang_Srinivasan_Kirchhoff_2025}               & 0.797 & 0.490 & 0.708 & 0.716 & 0.856 & 0.570 & 0.690 \\
% PromptAgent*~\cite{wang2024promptagent}     & 0.771 & 0.550 & 0.560 & 0.860* & 0.892 & 0.610 & 0.707 \\
PromptAgent~\cite{wang2024promptagent}     & 0.853 & \textbf{0.577} & 0.740 & 0.860* & 0.902 & 0.670 & 0.767 \\
\midrule
ZERA (Ours)     & \textbf{0.877} & 0.520 & \textbf{0.940} & \textbf{0.930} & \textbf{0.951} & \textbf{0.690} & \textbf{0.818} \\
\bottomrule
\end{tabular}
\vspace{-5pt}
\end{table*}

% 다른 요소들 언급이 안됨, 테이블 캡션에 해당 내용 언급, APE메소드는 citation달기

\subsection{Prompt Refinement from Zero Initialization}

ZERA is initialized with a deliberately underspecified prompt configuration, using a generic system prompt (\texttt{"You are a helpful assistant"}) and a minimal user prompt (\texttt{"Hello! I'm here to help you"}). Unlike prior approaches, ZERA does not rely on task-specific evaluation metrics. Instead, it leverages a multi-principle scoring framework grounded in generalizable, meta-level principles. Through iterative evaluation and refinement, ZERA progressively discovers prompts that guide the LLM toward outputs aligned with target responses. Notably, all experiments are conducted without access to task-specific knowledge—such as evaluation metrics or pre-defined task descriptions (often provided in datasets)—beyond a few example (5-20) instances drawn from the training data. Note that the “pre-defined task descriptions” mentioned here refer to those provided in benchmark datasets, and should not be confused with the task descriptions used earlier in this work, which are generated and refined as part of the optimization process.

% ZERA begins with a deliberately underspecified prompt configuration—setting the system prompt to \texttt{"You are a helpful assistant"} and the user prompt to \texttt{"Hello! I'm here to help you"}. In addition, ZERA does not require a primary evaluation metric that each task benchmark uses, by adapting multi-principle based scoring method. The only need is task example including desired outputs. ZERA iteratively test prompts on task examples and identify an optimalized prompt that makes LLM provide a similar output from the desried one. 

% \input{sections/3_2_prompt_composition_and_iterative_ptimization}

\section{Experiments}
\label{sec:experiments}

\subsection{APO Experimental Setting}

APO seeks to generate a task-specific prompt that enables a LLM to perform well on a given task, using only a small number (5-20) of representative samples. In this setting, the optimization process must rely on limited data while ensuring generalization across unseen examples. 

To simulate this scenario, we construct a task sample pool using the training and validation sets from standard benchmark datasets. The optimized prompt is then evaluated on the benchmark’s held-out test set using the official evaluation metrics defined for each task. This experimental protocol aligns with widely adopted practices in prior APO literature, ensuring consistency and comparability across different methods.

% \begin{adjustwidth}{0pt}{0pt}
% We evaluate ZERA in a two-stage experimental setup. First, we test its ability to iteratively generate high-quality prompts across a diverse set of NLP tasks (Section~4.1). Second, we assess how well these prompts transfer across different language models (Section~4.2). We also conduct ablation studies (Section~4.3) to isolate the contribution of key components—such as dynamic weighting and multi-criteria evaluation—to the overall refinement process.

\begin{adjustwidth}{0pt}{0pt}
Our benchmark suite spans nine datasets covering structured, unstructured, and reasoning-intensive tasks: GSM8K~\cite{cobbe2021training}, MMLU-Pro~\cite{hendrycks2021measuring}, and BBH~\cite{suzgun2022challenging} require symbolic or multi-step reasoning; MBPP~\cite{austin2021program} and HumanEval~\cite{austin2021program} involve functional code generation; CNN/DailyMail~\cite{hermann2015teaching}, SAMSum~\cite{gliwa2019samsum}, and HellaSwag~\cite{zellers2019hellaswag} test summarization and commonsense inference; and MMLU~\cite{hendrycks2021measuring} covers broad-domain factual QA. This diversity enables a comprehensive evaluation of ZERA’s prompt generalization capabilities across varying tasks.
\end{adjustwidth}

\subsection{Performance Comparison from Baselines}

To demonstrate the effectiveness of ZERA, we conducted a series of comparative experiments against state-of-the-art prompt optimization methods, including PromptAgent, OPRO, and CriSPO. To ensure fairness, each comparison was carried out under the original experimental settings proposed and reproduced by the respective methods. Specifically, we report results from (1) direct comparisons with OPRO and CriSPO, (2) head-to-head evaluation with PromptAgent, and (3) performance analysis across nine benchmark datasets, where ZERA is also compared against the default prompts provided by each benchmark. All experiments are conducted using a variety of LLMs to measure robustness and generalization across models and tasks.

\subsubsection{Comparison with OPRO and CriSPO}
\label{sec:opro-crispo-comparison}

We compare ZERA with two recent APO baselines, OPRO~\cite{yang2024large} and CriSPO~\cite{He_Liu_Xu_Shivade_Zhang_Srinivasan_Kirchhoff_2025}, on three tasks spanning math reasoning and summarization: GSM8K, CNN/DailyMail, and SAMSum. Following the original CriSPO setup, we evaluate all methods on 500 randomly sampled test instances per dataset using the LLaMA-3.1. Results for OPRO and CriSPO are reproduced using their official codebase.\footnote{\url{https://github.com/amazon-science/CriSPO}}

\begin{table}[t!]
\centering
\caption{GSM8K accuracy, CNN/DailyMail and SAMSum ROUGE-L scores evaluated with LLaMA-3.1.}
\vspace{-2pt}
\small
\setlength{\tabcolsep}{5pt}
\begin{tabular}{lccc|c}
\toprule
Method & GSM8K & CNN & Samsum & Avg. \\
\midrule
Baseline {\tiny (0 shot)} & 0.341 & 0.280 & 0.266 & 0.296 \\ 
Baseline {\tiny (5 shot)} & 0.357 & 0.296 & 0.286 & 0.313 \\
OPRO~\shortcite{yang2024large}         & 0.892 & 0.295 & 0.273 & 0.487 \\ 
CRiSPO~\shortcite{He_Liu_Xu_Shivade_Zhang_Srinivasan_Kirchhoff_2025}        & 0.896 & \textbf{0.309} & 0.270 & 0.492 \\ 
\midrule
ZERA  & \textbf{0.927} & 0.296 & \textbf{0.333} & \textbf{0.519} \\ 
\bottomrule
\end{tabular}
\vspace{-5pt}
\label{tab:comparison-opro-crispo-zera}
\end{table}

% 베이스라인 추가
% OPRO~\cite{yang2024large} and CriSPO~\cite{He_Liu_Xu_Shivade_Zhang_Srinivasan_Kirchhoff_2025}

As shown in Table~\ref{tab:comparison-opro-crispo-zera}, ZERA achieves the highest average performance across the three tasks, outperforming both OPRO and CriSPO on GSM8K and SAMSum. Notably, ZERA delivers a substantial improvement of +6.0 ROUGE-L on SAMSum, demonstrating strong capabilities in dialogue-style summarization. Appendix A2 shows the final prompt from ZERA in this task. 
%For the reproducible prompt configuration used by ZERA, see Appendix XX. 
%~\ref{tab:prompt-bbh-epistemic}, ~\ref{tab:prompt-samsum}.

\renewcommand{\arraystretch}{1.3}
\begin{table*}[t]
\centering
\caption{Performance comparison between baseline prompts and ZERA prompts across models and tasks. Each cell shows Baseline / ZERA score using the task’s standard evaluation metric. All values are reported as Baseline / ZERA score. EM = exact match, ROUGE-L = recall-oriented summary metric, pass@1 = functionally correct code generation on first attempt.}
\label{tab:full-results}
\vspace{-3pt}
% \resizebox{\textwidth}{!}{
\small
\begin{tabular}{l|ccccc}
\toprule
Dataset (Metric) & GPT-4o & GPT-3.5-turbo & LLaMA-3.1 & Qwen2.5 & Mistral-7B \\
\midrule
MMLU (EM) & 84.1 / \textbf{85.5} & 65.4 / \textbf{66.9} & \textbf{75.8} / 75.4 & \textbf{80.4} / 79.8 & \textbf{56.4} / 55.7  \\
MMLU-Pro (EM) & 58.7 / \textbf{75.3} & 37.3 / \textbf{46.2} & 50.8 / \textbf{60.1} & 54.5 / \textbf{72.8} & 30.0 / \textbf{30.1} \\
GSM8K (EM) & \textbf{95.8} / 95.3 & 72.55 / \textbf{78.2} & 34.1 / \textbf{92.6} & 92.12 / \textbf{96.1} & 11.5 / \textbf{53.0} \\
MBPP (pass@1) & 28.4 / \textbf{61.8} & 36.2 / \textbf{60.4} & 62.3 / \textbf{63.4} & 22.1 / \textbf{68.0} & 42.6 / \textbf{45.4} \\
HumanEval (pass@1) & 82.9 / \textbf{85.4} & \textbf{65.2} / 61.6 & 71.3 / \textbf{73.8} & 75.0 / \textbf{76.2} & 15.24 / \textbf{29.9} \\
BBH (EM) & 75.4 / \textbf{84.1} & 45.9 / \textbf{59.8} & 58.7 / \textbf{72.9} & 62.3 / \textbf{77.4} & 34.5 / \textbf{36.2} \\
HellaSwag (EM) & \textbf{90.6} / 90.0 & 46.3 / \textbf{66.6} & 81.6 / \textbf{84.2} & 87.8 / \textbf{89.2} & \textbf{66.0} / 62.6 \\
CNN/DM (ROUGE-L) & 27.8 / \textbf{29.0} & 28 / \textbf{29.9} & 28 / \textbf{29.6} & 26.5 / \textbf{30.0} & 28.0 / \textbf{29.8} \\
Samsum (ROUGE-L) & 27.7 / \textbf{38.2} & 28.0 / \textbf{31.9} & 26.2 / \textbf{33.7} & 29.8 / \textbf{36.0} & 24.5 / \textbf{34.0} \\

\midrule
Avg. Gain ($\Delta$) & \textbf{+8.1} & \textbf{+8.5} & \textbf{+10.8} & \textbf{+10.6} & \textbf{+7.6} \\
\bottomrule
\end{tabular}
% }
\vspace{-5pt}
% \small\textit{All values are reported as Baseline / ZERA score. EM = exact match, ROUGE-L = recall-oriented summary metric, pass@1 = functionally correct code generation on first attempt.}
\end{table*}

\subsubsection{Comparison with PromptAgent}
\label{sec:promptagent-comparison}

To further assess ZERA’s reasoning capabilities, we evaluate it against PromptAgent on six BBH sub-tasks—Penguins in a Table, Geometry, Epistemic Reasoning, Object Counting, Temporal Sequences, and Causal Judgment—following the experimental setup of the original PromptAgent paper. All evaluations are conducted using GPT-3.5-turbo as the base model for response generation, with GPT-4o used as the optimizer for prompt refinement in both ZERA and PromptAgent.

% 참고한 프롬프트 논문들 찾는다

As shown in Table~\ref{tab:bbh-subtasks}, ZERA outperforms PromptAgent in 5 out of 6 sub-tasks, including substantial gains in epistemic reasoning (+20.0) and temporal reasoning (+4.9). ZERA also achieves the highest overall average score (0.818), surpassing PromptAgent’s 0.767. These results highlight ZERA’s robust capabilities in complex multi-step reasoning and deep inference. Appendix A3 shows the final prompt from ZERA for the epistemic task.

%A representative example of the ZERA-generated prompt for the epistemic task is provided in Appendix~\ref{tab:prompt-bbh-epistemic}.

\subsubsection{Comparison with Primary Prompts}
\label{sec:general-comparison}
To evaluate ZERA’s generalization across model families and task types, we benchmark it using five diverse LLMs: GPT-3.5-turbo\cite{ye2023comprehensive}, GPT-4o\cite{openai2024gpt4o}, Qwen2.5-72B-Instruct\cite{qwen2024qwen25}, LLaMA-3.1-70B-Instruct\cite{dubey2024llama3}, and Mistral-7B-Instruct-v0.3\cite{jiang2023mistral}. All models are evaluated via API or open checkpoints without additional fine-tuning. We use the same nine benchmark datasets introduced in Section \ref{sec:experiments}, sampling 500 test instances per task, following the evaluation protocol of previous literature~\cite{He_Liu_Xu_Shivade_Zhang_Srinivasan_Kirchhoff_2025,wang2024promptagent}.

For baseline comparison, we adopt minimal yet format-compliant prompts (See Appendix A4.) that satisfy basic evaluation criteria without manual optimization. These serve as practical lower bounds for fair and reproducible measurement. Performance is measured using task-specific metrics: exact match for reasoning and classification tasks (e.g., MMLU, BBH, GSM8K), ROUGE-L for summarization (CNN/DailyMail, SAMSum), and pass@1 for code generation tasks (MBPP, HumanEval).

%\paragraph{Consistent gains across reasoning and generation tasks.}
ZERA consistently improves over baseline prompts across a variety of models and tasks (Table~\ref{tab:full-results}). The gains are especially pronounced on structured reasoning benchmarks: on GSM8K, for example, ZERA boosts LLaMA-3.1 to 92.6\% accuracy—approaching the 95.1\% reported in the original LLaMA paper using 8-shot chain-of-thought prompting~\citet{dubey2024llama3}. It also outperforms instruction-tuned models such as Qwen2.5, exceeding their published scores on GSM8K (91.5\% vs. 96.1\%) and MMLU-Pro (58.1\% vs. 72.8\%)~\citet{qwen2024qwen25}. These results highlight ZERA’s robustness across diverse models and tasks, even relative to expert-tuned few-shot configurations.

% \paragraph{Generalization and capacity-related limitations.}
% Beyond reasoning tasks, ZERA yields gains on open-ended generation such as CNN and SAMSum summarization, where iterative refinement induces abstraction-oriented prompt constraints (e.g., \emph{summarize concisely}, \emph{highlight key facts}) that improve ROUGE-L scores. These constraints emerge from evaluation-aware guidance, emphasizing \texttt{conciseness}, \texttt{meaning accuracy}, and \texttt{faithfulness}.

\subsubsection{APO Efficiency Comparison}

\begin{table}[ht!]
    \caption{Inference cost (\# of request / \# of tokens) comparison across OPRO, CriSPO and ZERA.}
    \label{tab:inference_cost_small}
    \centering
    \small
    \begin{tabular}{l|ccc|c}
    \toprule
    Method  & GSM8K & CNN & SAMSum & Avg. \\
    \midrule
    \multirow{2}{*}{OPRO}
    & 5,065/  & 15,024/   & 1,607/ & 7,232/ \\
    & 1,767K & 19,913K & \textbf{482K} & 7,387K \\
    \midrule
    \multirow{2}{*}{CriSPO}
    & 2,273/  & 1,509/   & 723/ & 1,504/ \\
    & 1,469K & 6,950K & 661K & 3,027K \\
    \midrule
    \multirow{2}{*}{ZERA}
    & \textbf{287}/  & \textbf{205}/   & \textbf{205}/ & \textbf{233}/ \\
    & \textbf{887K} & \textbf{759K} & 596K & \textbf{747K} \\
    \bottomrule
    \end{tabular}
\end{table}

The APO process diagnoses and improves task prompts based on multiple LLM calls. The number of LLM calls and tokens processed during the optimization of a single task prompt indicates the cost involved in prompt optimization. Table~\ref{tab:inference_cost_small} compares the costs required for APO across three benchmarks. As shown, ZERA demonstrates the lowest number of API calls due to its principle-based evaluation and improvement approach, thereby enabling APO with relatively fewer tokens processed.

\begin{figure}[t!]
\centering
\begin{tikzpicture}
\begin{axis}[
    width=6cm,
    height=4cm,
    xlabel={Number of samples},
    ylabel={Accuracy (\%)},
    xtick={5, 20, 50, 100, 200},
    ymin=70, ymax=100,
    grid=major,
    legend style={at={(0.5,1.35)}, anchor=north, legend columns=3},
    mark options={scale=1.2}
]

% ZERA (line)
\addplot[
    color=blue,
    thick,
    mark=*,
    mark options={fill=blue},
] coordinates {
    (5, 87.80)
    (20, 91.60)
    (50, 92.60)
    (100, 92.60)
    (200, 93.00)
};
\addlegendentry{ZERA}

% CRiSPO (dots only)
\addplot[
    color=red,
    % only marks,
    mark=square*,
    mark options={fill=red}
] coordinates {
    (5, 74.40)
    (20, 88.4)
    (50, 87.8)
    (100, 88.6)
    (200, 89.6)
};
\addlegendentry{CRiSPO}

% OPRO (dots only)
\addplot[
    color=green!60!black,
    % only marks,
    mark=square*,
    mark options={fill=green!60!black}
] coordinates {
    (5, 82.8)
    (20, 85.8)
    (50, 89)
    (100, 87.2)
    (200, 86.6)
};
\addlegendentry{OPRO}

\end{axis}
\end{tikzpicture}
\caption{APO performance comparison across varied task sample sizes, required to conduct the prompt optimization for GSM8K. The comparison shows how many task samples are required to identify optimized the task prompt through APO methods.}
\label{fig:sample_ablation}
\end{figure}
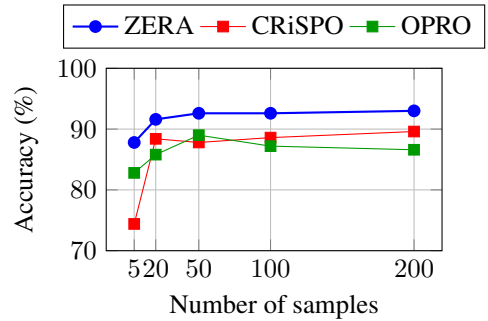

Additionally, the number of task samples required for the APO process is a critical resource, as creating samples to define a task is highly cost-intensive. Figure~\ref{fig:sample_ablation} illustrates a performance comparison based on the size of task samples utilized during APO. Despite defining and utilizing only 20 task samples, ZERA achieves higher performance than CRiSPO and OPRO, which rely on 200 samples—10 times the quantity. As demonstrated, ZERA can attain a high level of APO with fewer samples, showcasing the efficacy of its principle-based prompt critique mechanism.

\subsection{Process Analysis of ZERA}
\label{sec:prompt-evolution}

% 4.2 

% 4.3
% 1. Process Analysis 
% - Zero prompt에서 출발하지만 보는 것처럼 좋은 prompt가 생성되는 것을 관측함
% 2. 적은 샘을 이용함에도 불구하고 converge 되고 그 단계가 a few iteration에서도 관측됨을 확인함
% - unified score가 converge는 오케이
% - task performance converge는 아님
% (주장하기 모호한 감이 있음)
% 3. Robustness (가능하다면)

% 4.4
% 4.4.1 - Ablation on principle weight
% * 벤치마크에 따라서 다른 principle weight를 가짐 (Table 11 축약해 보기)
% 4.4.2 - Ablation on principles
% 4.4.3 - Ablation on task prompt structure
% 4.4.4 - Prompt Transferability

% 그외 (어펜딕스)
% Table 12-14 중 1개 선정해서 본문으로 이동 + 남은것은 어펜딕스
% Table 17은 (base prompt) 에펜딕스 남기기
% Table 20은 (principle 설명) 살리기 
% Table 18, 19 제거 (경우에 따라서 살릴 수 있을 것) - 구성 봐서

% Table 15,16은 (dynamic ablation)은 제거 -- 내용을 본문에 살짝 담기만 하고
% Table 9. Prompt에서 일부 수정한 것 - 이것 없애자 + Baseline은 위의 OPRO Table하고 합치기
% Table 10. 지우자

As described in Section 3.5, ZERA begins with zero prompt initialization and optimizes based solely on a small number of task samples—typically around five per iteration. In this section, we analyze ZERA's optimization dynamics from two perspectives: (1) tracking the trajectory of the unified evaluation score to illustrate how the prompt converges over iterations, and (2) qualitatively examining how the prompt content evolves and expands throughout the refinement process.

\subsubsection{Analysis on Evaluation Score Trajectory}

% ZERA begins with a deliberately underspecified prompt configuration—setting the system prompt to \texttt{"You are a helpful assistant"} and the user prompt to \texttt{"Hello! I'm here to help you"}. From this minimal baseline, it iteratively transforms prompts into structurally precise, evaluation-aligned formats. 
% Prompt refinement is performed separately for each dataset using GPT-3.5-turbo~\cite{ye2023comprehensive}, leveraging a small randomly sampled training subset of 5–20 examples (typically only 5 are sufficient). The refinement process relies exclusively on the training and validation splits, while final evaluation is conducted on the held-out test set to ensure proper generalization.

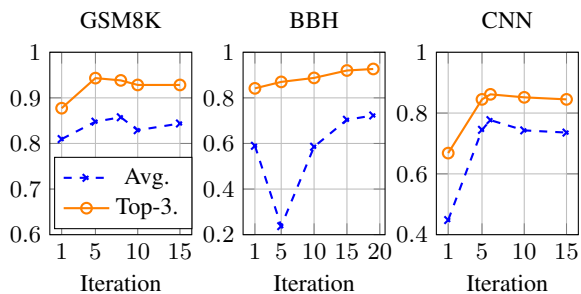
\begin{figure}[t!]
\centering

\footnotesize
% --- GSM8K ---
\begin{subfigure}{0.32\linewidth}
\centering
\begin{tikzpicture}
\begin{axis}[
    title={GSM8K},
    xlabel={Iteration},
    xtick={1,5,10,15},
    ymin=0.6, ymax=1,
    width=1.4\textwidth,
    height=4cm,
    grid=both,
    legend pos=south west,
]
\addplot+[
    mark=x, thick, blue, dashed,
    error bars/.cd,
    y dir=both,
    y explicit
] coordinates {
    (1,0.8096)
    (5,0.8478) 
    (8,0.8575) 
    (10,0.8291)
    (15,0.8438)
};
\addlegendentry{Avg.}

\addplot+[mark=o, thick, orange] coordinates {
    (1,0.8773) (5,0.9433) (8,0.9383) (10,0.9283) (15,0.9283)
};
\addlegendentry{Top-3.}

\end{axis}
\end{tikzpicture}
\end{subfigure}
%
% --- MMLU-Pro ---
\begin{subfigure}{0.32\linewidth}
\centering
\begin{tikzpicture}
\begin{axis}[
    title={BBH},
    xlabel={Iteration},
    ymin=0.2, ymax=1,
    xtick={1,5,10,15,20},
    width=1.4\textwidth,
    height=4cm,
    grid=both,
]
% Avg Score 
\addplot+[
    mark=x, thick, blue, dashed,
    error bars/.cd,
    y dir=both,
    y explicit
] coordinates {
    (1,0.5898) 
    (5,0.2373) 
    (10,0.5856)
    (15,0.7039)
    (19,0.7228) 
};
% \addlegendentry{Avg Score}

% Top-3 Avg (Orange)
\addplot+[mark=o, thick, orange] coordinates {
    (1,0.8417) (5,0.8700) (10,0.8875) (15,0.9200) (19,0.9267)
};
% \addlegendentry{Top-3 Avg}

\end{axis}
\end{tikzpicture}
\end{subfigure}
%
% --- CNN ---
\begin{subfigure}{0.32\linewidth}
\centering
\begin{tikzpicture}
\begin{axis}[
    title={CNN},
    xlabel={Iteration},
    xtick={1,5,10,15},
    ymin=0.4, ymax=1,
    width=1.4\textwidth,
    height=4cm,
    grid=both
]
\addplot+[
    mark=x, thick, blue, dashed,
    error bars/.cd,
    y dir=both,
    y explicit
] coordinates {
    (1,0.4480)
    (5,0.7448)
    (6,0.7767)
    (10,0.7430)
    (15,0.7358)
};

\addplot+[mark=o, thick, orange] coordinates {
    (1,0.6683) (5,0.8450) (6,0.8616) (10,0.8517) (15,0.8450)
};

\end{axis}
\end{tikzpicture}
\end{subfigure}
\vspace{-2pt}
\caption{
The trajectories of evaluation scores identified by PCG. Each iteration samples 5 task examples and evaluate the current prompt based on the eight principles. Avg. anc Top-3. indicate the average over all sampled examples and the average of top-3 scored samples.
}
\label{fig:zera-iteration-errorbars}
\vspace{-5pt}
\end{figure}

We analyze how prompt quality evolves over refinement iterations by tracking the unified evaluation score at each step (up to 20 iterations). Figure~\ref{fig:zera-iteration-errorbars} shows the score trajectories for three representative datasets: (GSM8K, BBH, and CNN). Substantial gains often emerge within the first 1–5 iterations, especially in \textsc{GSM8K} and \textsc{CNN}, which tend to converge quickly with as few as 5 training examples. In contrast, \textsc{BBH}, which requires more complex reasoning, show continued improvement even in later iterations, reflecting the benefit of extended refinement on more complex task structures. 

Although each iteration of ZERA uses only a small number of task samples, we observe that the resulting prompts yield stable unified scores across steps. This indicates that the principle-based evaluation and prompt refinement process remains stable, even as the task samples vary at each step. These findings suggest that ZERA’s optimization trajectory is both stable and convergent, with minimal fluctuation in performance despite changes in the evaluation data per iteration.

\subsubsection{Analysis on Prompt Evolution}

% \subsubsection{Prompt Structure Evolution}

ZERA incrementally transforms underspecified prompts into task-adapted formats through iterative self-refinement. Across iterations, the prompts increasingly encode task structure, role assignments, output constraints, and formatting conventions---progressively aligning with task-specific demands. This evolution occurs both semantically (e.g., shifting from vague to expert roles) and structurally (e.g., introducing reasoning steps or enforcing output schemas).

% This transformation is governed by the MPR, which leverages refinement history to generate updated task representations and prompt candidates. As formalized in Equation~\ref{eq:history}, the agent maps accumulated prompt-output  into structurally aligned prompts by internalizing feedback from multi-dimensional evaluation over time. This iterative adaptation enables ZERA to progressively refine prompts in response to emerging task patterns and scoring signals.

\begin{table}[ht!]
\centering
\caption{Prompt evolution across iterations on BBH.}
\label{tab:prompt-evolution-bbh}
\footnotesize
\setlength{\tabcolsep}{3pt}
\begin{tabular}{p{0.3cm}|p{7cm}|}
\toprule
\#  & System Prompt \\
\midrule
1 & You are a helpful assistant. \\
2 & You are a helpful AI assistant. Reason freely through problems before providing precise, concise responses formatted clearly per the question's requirements. \\
19 & \textbf{You are a logical reasoning expert. Clearly reason each question step-by-step in natural, explicit language. Upon completing your analysis, distinctly separate it from your final concise answer, which must strictly follow the provided formatting instructions.} \\
\bottomrule
\end{tabular}
% \vspace{0.5em}
\begin{tabular}{p{0.3cm}|p{7cm}|}
\toprule
\# & User Prompt \\
\midrule
1 & Hello! I'm here to help you. \\
2 & Please answer the following questions clearly and concisely. [ZERA-generated reasoning exemplar, 1-shot] Begin now. \\
19 & \textbf{Solve these logical reasoning problems by explicitly thinking through them step-by-step before providing your final answer.[ZERA-generated reasoning exemplar, 3-shot] Now, begin solving.} \\
\bottomrule
\end{tabular}
% \par\smallskip
% \small\textit{Shown: Prompt refinement example from the \textbf{BBH} dataset. ZERA progressively induces structured, evaluation-aligned formats, including self-generated reasoning exemplars (e.g., 3-shot) guided by task feedback. Additional examples for \textbf{SAMSum} and \textbf{GSM8K} are provided in Tables~\ref{tab:prompt-evolution-samsum} and~\ref{tab:prompt-evolution-gsm8k} in the Appendix.}
\vspace{-10pt}
\end{table}

As shown in Table~\ref{tab:prompt-evolution-bbh}, ZERA adaptively introduces self-generated reasoning exemplars and reasoning scaffolds for BBH, adopts a question → reasoning → answer format for BBH. These structures emerge not from handcrafted examples, but through self-refinement using task-weighted feedback. These evolved prompts converge toward task-effective formats without relying on external supervision or manual prompt engineering. More prompt optimization results on other benchmarks can be found in Appendix A5.

%(Table~\ref{tab:adaptive-weights})

% \begin{table*}[h]
% \centering
% \caption{
% ZERA Refinement Results on 3 Benchmarks (Iter 1–15).  
% This table summarizes the performance trends across iterations, including the best-performing iteration for each dataset. See full results in Table~\ref{tab:zera-iterations}.
% }
% \label{tab:zera-iterations-concise}
% \small
% \begin{tabular}{l l >{\bfseries}c c c c c }
% \toprule
% \textbf{Dataset} & \textbf{Metric} & \textbf{Best Avg.} & \textbf{Iter 1} & \textbf{Iter 5} & \textbf{Iter 10} & \textbf{Iter 15} \\
% \midrule
% \multirow{3}{*}{\makecell{\textbf{MMLU-Pro} \\ \footnotesize (Best Iter:12)}} 
%   & Top-3 Avg & 0.9100 & 0.9067 & 0.8783 & 0.8567 & 0.8767 \\
%   & Avg Score & 0.7151 & 0.5868 & 0.6383 & 0.5958 & 0.5878 \\
%   & Std Dev   & 0.2113 & 0.2751 & 0.2183 & 0.2176 & 0.2422 \\
% \midrule
% \multirow{3}{*}{\makecell{\textbf{GSM8K} \\ \footnotesize (Best Iter:8)}} 
%   & Top-3 Avg & 0.9383 & 0.8773 & 0.9433 & 0.9283 & 0.9283 \\
%   & Avg Score & 0.8575 & 0.8096 & 0.8478 & 0.8291 & 0.8438 \\
%   & Std Dev   & 0.1203 & 0.0878 & 0.1559 & 0.1335 & 0.0962 \\
% \midrule
% \multirow{3}{*}{\makecell{\textbf{CNN} \\ \footnotesize (Best Iter:6)}} 
%   & Top-3 Avg & 0.8616 & 0.6683 & 0.8450 & 0.8517 & 0.8450 \\
%   & Avg Score & 0.7767 & 0.4480 & 0.7448 & 0.7430 & 0.7358 \\
%   & Std Dev   & 0.0555 & 0.1988 & 0.0592 & 0.0822 & 0.0723 \\
% \midrule
% \end{tabular}
% \end{table*}
% \input{sections/4_2_0_evaluation_result}
\subsection{Ablation Studies}

% Skeleton
% 4.4
% 4.4.1 - Ablation on principles
% 4.4.1 - Ablation on principle weight
% * 벤치마크에 따라서 다른 principle weight를 가짐 (Table 11 축약해 보기)

% 4.4.3 - Ablation on task prompt structure
% 4.4.4 - Prompt Transferability

Beyond overall performance, we conduct a focused ablation study to assess the contribution of key components in ZERA, including its scoring strategy, evaluation criteria, prompt component coverage, and base model alignment.%

% \begin{table}[t!]
% \centering
% \caption{
% Effect of evaluation criteria subsets on ZERA performance using \texttt{GPT-3.5-turbo}. 
% \textbf{8 dims.} uses all eight evaluation criteria; 
% \textbf{2 dims.} uses only \textit{correctness} and \textit{reasoning quality}; 
% \textbf{6 dims.} uses the remaining six dimensions (excluding the two above). 
% ZERA performs best when all eight criteria are used.
% }
% \label{tab:criteria-ablation}
% \small
% \begin{tabular}{l|c|c|c|c}
% \toprule
% \textbf{Dataset} & \textbf{8 dims.} & \textbf{2 dims.} & \textbf{6 dims.} & \textbf{Baseline} \\
% \midrule
% BBH & \textbf{59.8} & 26.2 & 55.2 & 45.9 \\
% MMLU-Pro & \textbf{46.2} & 45.4 & 43.2 & 37.3 \\
% \bottomrule
% \end{tabular}
% \end{table}

\begin{figure}[t!]
\vspace{-2pt}
\centering
\includegraphics[width=0.98\linewidth]{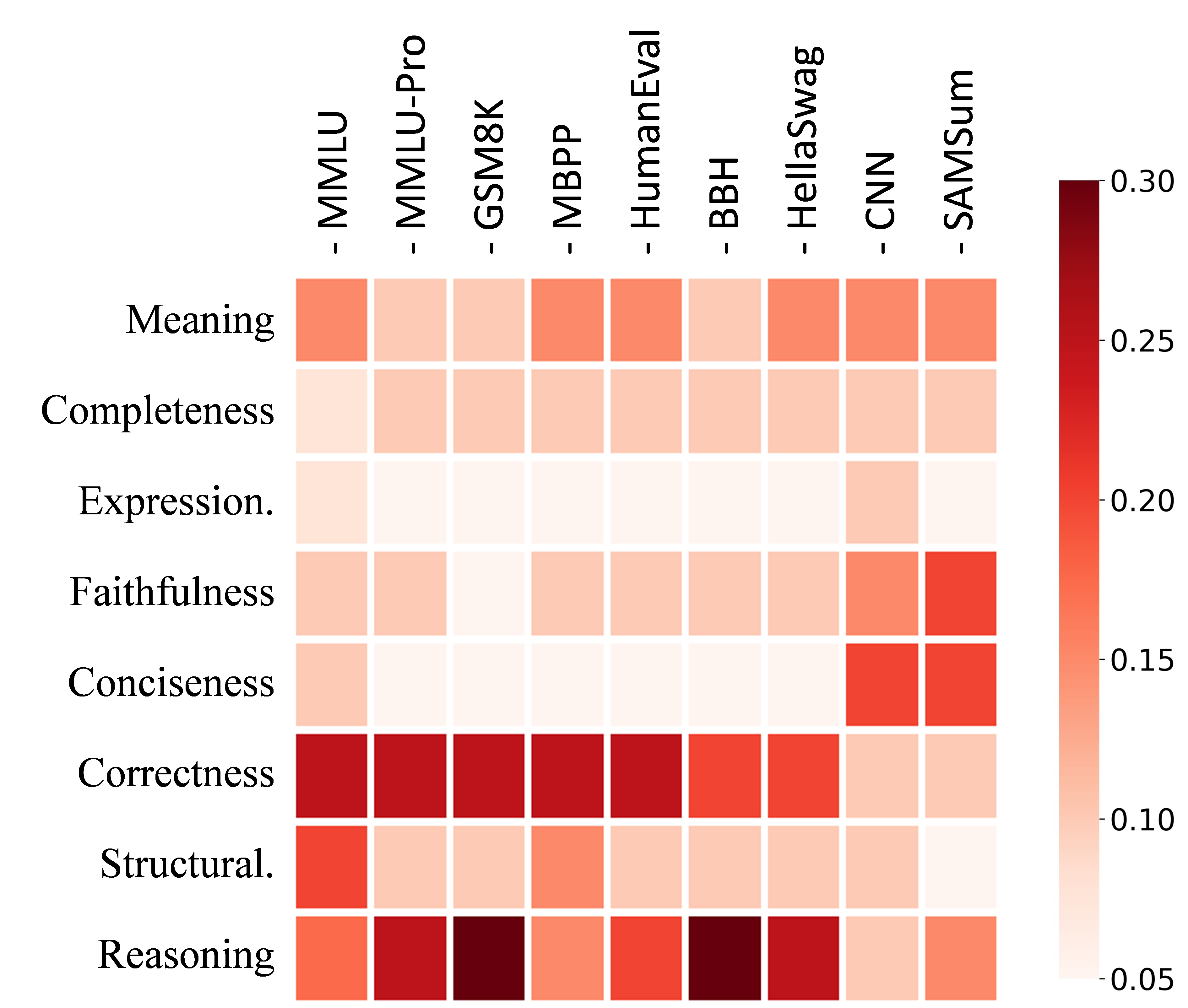}
\caption{Visualization of task-adaptive scoring weights over nine benchmarks. The values are averaged over task examples, sampled at the optimal step from the experiment in section 4.2.3.}
\label{fig:weight-figure}
\vspace{-5pt}
\end{figure}

\begin{table}[t]
\centering
\caption{ Ablation on task-adaptive principle weight. Fixed. indicates to ZERA using uniform weights; Dynmaic. refers to ZERA with task-adaptive weights.}
\vspace{-2pt}
\small
\label{tab:weight-ablation}
\begin{tabular}{l|cc}
\toprule
principle weight type & BBH & MMLU-Pro \\
\midrule
Fixed. (uniform) & 42.6 & 41.1 \\
Dynamic. & \textbf{59.8} & \textbf{46.2} \\
\bottomrule
\end{tabular}
\vspace{-5pt}
\end{table}

\subsubsection{Analysis on Principle Weights}

% \paragraph{Impact of Task-Adaptive Scoring.}
Beyond the structure and content of the prompts themselves, the evaluation mechanism used during refinement plays a critical role in overall performance. To assess this, we compare three variants: a minimal baseline, ZERA with fixed uniform weights, and full ZERA with dynamically inferred task-specific weights. As shown in Table~\ref{tab:weight-ablation}, dynamic weighting consistently improves performance in BBH and MMLU-Pro, validating the effectiveness of task-adaptive prioritization. The fixed-weight variant generally performs between the baseline and full ZERA, indicating that structure-inducing refinement offers meaningful benefits, while task-specific weighting further amplifies these gains.

We further investigate how the principle weights vary across different types of tasks, shown in Figure~\ref{fig:weight-figure}. 
%Table~\ref{tab:adaptive-weights} summarizes these task-adaptive weights (scaled to 0–1) across datasets. 
They guide MPG toward structure-sensitive prompt strategies tailored to each task’s demands. For instance, ``reasoning quality'' receives the highest weight in tasks such as GSM8K and MMLU-Pro, both of which demand multi-step logical inference. Meanwhile, ``correctness'' is also emphasized in MMLU-Pro and MMLU, reflecting its need for factual precision in knowledge-intensive QA. In contrast, summarization tasks like CNN and SAMSum assign greater weight to ``conciseness'' and ``faithfulness'', highlighting the importance of generating informative yet succinct summaries.

These task-adaptive scoring patterns indicate that PCG aligns evaluation emphasis with task demands—prioritizing structural, semantic, or reasoning criteria as needed—without relying on manual heuristics or fixed weights.

\begin{table}[t!]
\centering
\caption{Effect of principle-based criteria. Baseline evaluates prompts without any principles and other utilize the subset or all principles.}
\label{tab:criteria-ablation}
\small
\begin{tabular}{l|cc}
\toprule
Criteria & BBH & MMLU-Pro \\
\midrule
No principles (baseline) & 45.9 & 37.3 \\
Correctness, reasoning & 26.2 & 45.4 \\ 
All w/o correctness, reasoning & 55.2 & 43.2 \\
All eight principles & \textbf{59.8} & \textbf{46.2} \\
\bottomrule
\end{tabular}
\vspace{-3pt}
\end{table}

\subsubsection{Analysis on Principles}

We investigate how the number and type of evaluation criteria affect prompt refinement. Specifically, we compare three variants of ZERA: one using all eight criteria, one using only two (\textit{reasoning quality} and \textit{correctness}), and one using the remaining six. This split reflects the high weight scores of \textit{Correctness} and \textit{Reasoning}, observed from Figure~\ref{fig:weight-figure}. As shown in Table~\ref{tab:criteria-ablation}, using the full set of eight criteria yields the best performance on both BBH and MMLU-Pro. Reducing the evaluation to only two dimensions leads to a substantial drop on BBH (–33.6), highlighting the importance of structural and stylistic signals in tasks requiring multi-step reasoning. Even when using six criteria, performance remains slightly below the full setting, suggesting that ZERA benefits from a holistic view of output quality that balances reasoning, faithfulness, clarity, and structure.

% \begin{table}[t]
% \centering
% \caption{
% Effect of evaluation strategy on prompt optimization using \texttt{GPT-3.5-turbo}. 
% \textbf{Dynamic.} refers to ZERA using task-adaptive weights; 
% \textbf{Fixed.} refers to ZERA using uniform weights (0.125) across criteria; 
% \textbf{Baseline} denotes the unrefined baseline prompt.
% }
% \small
% \label{tab:weight-ablation}
% \begin{tabular}{l|ccc}
% \toprule
% \textbf{Dataset} & \textbf{Dynamic.} & \textbf{Fixed.} & \textbf{Baseline} \\
% \midrule
% BBH & \textbf{59.8} & 42.6 & 45.9  \\
% MMLU-Pro & \textbf{46.2} & 41.1 & 37.3 \\
% \bottomrule
% \end{tabular}
% \end{table}

\begin{table}[t!]
\centering
\caption{Ablation study on the prompt components. User Only indicates no use of system prompt in a targeted task prompt.}
%{GSM8K accuracy, CNN/DailyMail and SAMSum ROUGE-L scores evaluated with LLaMA 3.1 70B for ZERA, ZERA w/o Task, and ZERA only User Prompt}
\small
% \tabcolsep
% \resizebox{0.95\linewidth}{!}{%
\setlength{\tabcolsep}{4pt}
\begin{tabular}{lcccc|c}
\toprule
Method & GSM8K & CNN & Samsum & BBH & Avg. \\
\midrule
w/o $\text{T}^{(t)}_{\text{task}}$         & \textbf{0.930} & 0.266 & \textbf{0.345} & 0.728 & 0.567 \\ 
User Only       & 0.914 & 0.270 & 0.327 & 0.726 & 0.559  \\ 
\midrule
ZERA  & 0.927 & \textbf{0.296} & 0.337 & \textbf{0.729} & \textbf{0.571} \\ 
\bottomrule
\end{tabular}
% }
\label{tab:comparison-zera-w/otask-onlyprompt}
\vspace{-5pt}
\end{table}

\subsubsection{Ablation on Prompt Components}
% \paragraph{Effect of Prompt Component Coverage.}
Complementing the analysis of evaluation criteria diversity, we examine how the structure of the prompt itself, specifically, the inclusion of different prompt components, affects performance. We compare the full version of ZERA, which incorporates the system prompt, task specification, and user prompt, with two ablated variants: one that omits the explicit task type definition (\textit{w/o Task}) and another that uses only the user prompt (\textit{User Only}). As shown in Table~\ref{tab:comparison-zera-w/otask-onlyprompt}, both variants result in performance drops across tasks, with the User Only setting yielding the lowest average score. These results suggest that including both task specification and system-level intent improves alignment with evaluation objectives and enables more effective prompt optimization.

% \begin{table}[t!]
% \centering
% \caption{Effect of Base Model Alignment on Prompt Transferability. Each cell reports accuracy when applying prompts tuned on GPT-3.5 or LLaMA to LLaMA-3.1.}
% \label{tab:base-model-alignment}
% \small
% \begin{tabular}{l|cc}
% \toprule
% \textbf{Dataset} & \makecell{\textbf{ZERA}} & \makecell{\textbf{ZERA}} \\
% \midrule
% APO Agent & GPT-3.5 & LLaMA \\
% LLM & LLaMA & LLaMA \\
% \midrule
% BBH      & 72.9 & \textbf{76.9} \\
% MMLU-Pro & 57.3 & \textbf{60.7} \\
% GSM8K    & 92.6 & \textbf{92.7} \\
% MBPP     & 57.9 & \textbf{58.3} \\
% \bottomrule
% \end{tabular}
% \vspace{0.5em}
% \begin{minipage}{0.9\linewidth}
% \small\textit{Prompts refined using different base models yield varying levels of transfer performance.}
% \end{minipage}
% \end{table}

\begin{table}[t!]
\centering
\caption{Analysis on transferability of optimized prompt by ZERA. $\text{LLM}_{\text{ZERA}}$ indicates LLM model used in both PCG and MPR. 
%LLaMA indicates LLaMA-3.1-70B-Instruct.
}
\vspace{-3pt}
\setlength{\tabcolsep}{2.9pt}
\label{tab:base-model-alignment}
\small
\begin{tabular}{cc|cccc}
\toprule
$\text{LLM}_{\text{ZERA}}$ &  $\text{LLM}_{task}$ & BBH & MMLU-Pro & GSM8K & MBPP \\
\midrule
GPT-3.5 & LLaMA & 72.9 & 57.3 & 92.6 & 57.9 \\
LLaMA & LLaMA & \textbf{76.9} & \textbf{60.7} & \textbf{92.7} & \textbf{58.3} \\
\bottomrule
\end{tabular}
% \vspace{0.5em}
% \begin{minipage}{0.9\linewidth}
% \small\textit{Prompts refined using different base models yield varying levels of transfer performance.}
% \end{minipage}
\vspace{-5pt}
\end{table}

\subsubsection{Analysis on Transferability of Prompt}
Lastly, we assess how the alignment between the base model used during prompt refinement and the model used at inference time affects performance. Specifically, we compare prompts refined using GPT-3.5-turbo and LLaMA-3.1, with both evaluated on LLaMA-3.1. As shown in Table~\ref{tab:base-model-alignment}, prompts optimized on LLaMA-3.1 consistently outperform those generated with GPT-3.5, across all tasks. The gap is most notable on BBH and MMLU-Pro, where alignment between the refinement-time and inference-time models appears crucial for maximizing performance. While prompts transferred from GPT-3.5 still yield competitive results (e.g., 92.6 on GSM8K), model-specific nuances—especially in reasoning or formatting—are better captured when prompts are tuned on the target architecture.

% \paragraph{Ablation Summary.} Across all ablation settings, ZERA demonstrates robust performance and clear structural advantages. Key components such as CoT cues, dynamic scoring, and model alignment all contribute to its effectiveness, validating the importance of structure-aware and model-adaptive refinement. (See Appendix~\ref{appendix:cot_vs_examples} for detailed component-wise results.)

\section{Conclusion}

This paper introduces ZERA, a novel APO method that operates solely on target task samples without relying on predefined initial prompt and evaluation metrics. ZERA generates critiques of prompt outputs based on eight generalizable principles and refines prompts accordingly through an iterative process. By leveraging prompt update history and principle-based scoring, ZERA achieves stable refinement and consistently converges toward high-performing prompts. Extensive experiments across diverse tasks and models demonstrate the efficiency and effectiveness of the proposed approach. These results highlight ZERA's potential as a general-purpose, model-agnostic solution for scalable and interpretable prompt engineering across a wide range of domains.

% \paragraph{Theory-Grounded Prompt Evolution.}
% ZERA refines prompts through a principled optimization objective rather than heuristics. At each iteration, it maximizes alignment between model outputs and multi-dimensional critique feedback. The Principle-based Critique dynamically adjusts task-specific weights across eight criteria, guiding prompt evolution without relying on fixed templates or handcrafted exemplars.

% \paragraph{Generalization and Modular Compatibility.}
% Though optimized on GPT-3.5-turbo, ZERA generalizes well across models such as LLaMA-3.1 and Qwen2.5, often improving even further with re-optimization. Its modular design—separating evaluation, meta-prompt generation, and base models—ensures long-term adaptability to future LLMs and scoring protocols.

% \paragraph{Toward Self-Improving Reasoning Systems.}
% ZERA demonstrates that effective prompt structures can emerge through iterative feedback. By internalizing prompt-output history, it evolves reasoning scaffolds and task-aligned formats autonomously. We position ZERA as a lightweight foundation for building scalable, self-improving systems that adapt across domains, tasks, and model families.
\section{Limitations}

While ZERA demonstrates strong performance across diverse tasks and models, it has several limitations. First, our score reporting on summarization tasks such as CNN/DailyMail relies entirely on automatic metrics (e.g., ROUGE-L) without human judgment, which may overlook nuances like coherence or factuality. Second, although ZERA operates with minimal supervision, it still requires a small number of training samples (typically 5–20) for each task. Fully zero-shot refinement remains an open challenge. Third, as prompts evolve over iterations, they often become longer to encode structural or reasoning constraints. While this improves accuracy, it may lead to increased inference latency or context overflow in constrained environments. However, we observe that optimized prompts typically converge to a stable length after the early refinement stages rather than growing indefinitely. Thus, this limitation highlights an area for further efficiency improvements rather than an impractical barrier to deployment. Lastly, ZERA depends on an internal LLM to provide multi-level criteria feedback. Although effective in practice, its reliability under ambiguous or adversarial outputs has not been fully analyzed and may introduce bias in certain edge cases.

\bibliography{custom}

\begin{thebibliography}{27}
\providecommand{\natexlab}[1]{#1}

\bibitem[{Austin et~al.(2021)Austin, Odena, Nye, Bosma, Michalewski, Dohan, Jiang, Cai, Terry, Le et~al.}]{austin2021program}
Jacob Austin, Augustus Odena, Maxwell Nye, Maarten Bosma, Henryk Michalewski, David Dohan, Ellen Jiang, Carrie Cai, Michael Terry, Quoc Le, and 1 others. 2021.
\newblock \href {https://arxiv.org/abs/2108.07732} {Program synthesis with large language models}.
\newblock \emph{arXiv preprint arXiv:2108.07732}.

\bibitem[{Brown et~al.(2020)Brown, Mann, Ryder et~al.}]{brown2020language}
Tom Brown, Benjamin Mann, Nick Ryder, and 1 others. 2020.
\newblock Language models are few-shot learners.
\newblock In \emph{NeurIPS}.

\bibitem[{Chen et~al.(2025)Chen, Wang, Jiang, and Nakashima}]{Chen_Wang_Jiang_Nakashima_2025}
Junhao Chen, Bowen Wang, Zhouqiang Jiang, and Yuta Nakashima. 2025.
\newblock \href {https://doi.org/10.1609/aaai.v39i22.34527} {Putting people in llms’ shoes: Generating better answers via question rewriter}.
\newblock \emph{Proceedings of the AAAI Conference on Artificial Intelligence}, 39(22):23577--23585.

\bibitem[{Chen et~al.(2024)Chen, Wen, Fan, Chen, Wu, Liu, Li, Liu, and Xiao}]{chen2024mapo}
Yuyan Chen, Zhihao Wen, Ge~Fan, Zhengyu Chen, Wei Wu, Dayiheng Liu, Zhixu Li, Bang Liu, and Yanghua Xiao. 2024.
\newblock Mapo: Boosting large language model performance with model-adaptive prompt optimization.
\newblock \emph{arXiv preprint arXiv:2407.04118}.

\bibitem[{Cobbe et~al.(2021)Cobbe, Kosaraju, Bavarian, Chen, Jun, Kaplan et~al.}]{cobbe2021training}
Karl Cobbe, Vineet Kosaraju, Mohammad Bavarian, Mark Chen, Heewoo Jun, Jared Kaplan, and 1 others. 2021.
\newblock \href {https://arxiv.org/abs/2110.14168} {Training verifiers to solve math word problems}.
\newblock \emph{arXiv preprint arXiv:2110.14168}.

\bibitem[{Dubey et~al.(2024)Dubey, Jauhri, Pandey, Kadian, Al-Dahle, Letman, Mathur, Schelten, Yang, Fan et~al.}]{dubey2024llama3}
Abhimanyu Dubey, Abhinav Jauhri, Abhinav Pandey, Abhishek Kadian, Ahmad Al-Dahle, Aiesha Letman, Akhil Mathur, Alan Schelten, Amy Yang, Angela Fan, and 1 others. 2024.
\newblock \href {https://arxiv.org/abs/2407.21783} {The llama 3 herd of models}.
\newblock \emph{arXiv preprint arXiv:2407.21783}.

\bibitem[{Gliwa et~al.(2019)Gliwa, Mochol, Biesek, and Wawer}]{gliwa2019samsum}
Bogdan Gliwa, Iwona Mochol, Michał Biesek, and Aleksander Wawer. 2019.
\newblock \href {https://arxiv.org/abs/1911.12237} {Samsum corpus: A human-annotated dialogue summary dataset}.
\newblock \emph{arXiv preprint arXiv:1911.12237}.

\bibitem[{He et~al.(2025)He, Liu, Xu, Shivade, Zhang, Srinivasan, and Kirchhoff}]{He_Liu_Xu_Shivade_Zhang_Srinivasan_Kirchhoff_2025}
Han He, Qianchu Liu, Lei Xu, Chaitanya Shivade, Yi~Zhang, Sundararajan Srinivasan, and Katrin Kirchhoff. 2025.
\newblock \href {https://doi.org/10.1609/aaai.v39i22.34575} {Crispo: Multi-aspect critique-suggestion-guided automatic prompt optimization for text generation}.
\newblock \emph{Proceedings of the AAAI Conference on Artificial Intelligence}, 39(22):24014--24022.

\bibitem[{Hendrycks et~al.(2021)Hendrycks, Burns, Kadavath, Arora, Basart, Tang, Song, Steinhardt et~al.}]{hendrycks2021measuring}
Dan Hendrycks, Collin Burns, Saurav Kadavath, Akul Arora, Steven Basart, Dawn Tang, Dawn Song, Jacob Steinhardt, and 1 others. 2021.
\newblock \href {https://arxiv.org/abs/2009.03300} {Measuring massive multitask language understanding}.
\newblock \emph{arXiv preprint arXiv:2009.03300}.

\bibitem[{Hermann et~al.(2015)Hermann, Kocisky, Grefenstette, Espeholt, Kay, Suleyman, and Blunsom}]{hermann2015teaching}
Karl~Moritz Hermann, Tom{\'a}s Kocisky, Edward Grefenstette, Lasse Espeholt, Will Kay, Mustafa Suleyman, and Phil Blunsom. 2015.
\newblock \href {https://arxiv.org/abs/1506.03340} {Teaching machines to read and comprehend}.
\newblock \emph{Advances in Neural Information Processing Systems}, 28.

\bibitem[{Jafari et~al.(2024)Jafari, Mekala, Yu, and Berg-Kirkpatrick}]{jafari-etal-2024-morl}
Yasaman Jafari, Dheeraj Mekala, Rose Yu, and Taylor Berg-Kirkpatrick. 2024.
\newblock \href {https://doi.org/10.18653/v1/2024.findings-emnlp.577} {{MORL}-prompt: An empirical analysis of multi-objective reinforcement learning for discrete prompt optimization}.
\newblock In \emph{Findings of the Association for Computational Linguistics: EMNLP 2024}, pages 9878--9889, Miami, Florida, USA. Association for Computational Linguistics.

\bibitem[{Jiang et~al.(2023)}]{jiang2023mistral}
Zhen Jiang and 1 others. 2023.
\newblock \href {https://arxiv.org/abs/2310.06825} {Mistral 7b}.
\newblock \emph{arXiv preprint arXiv:2310.06825}.

\bibitem[{OpenAI(2024)}]{openai2024gpt4o}
OpenAI. 2024.
\newblock \href {https://arxiv.org/abs/2410.21276} {Gpt-4o system card}.
\newblock \emph{arXiv preprint arXiv:2410.21276}.

\bibitem[{Peng et~al.(2025)Peng, Zhou, Chen, Liu, Chen, and Qin}]{peng2025dlpo}
Dengyun Peng, Yuhang Zhou, Qiguang Chen, Jinhao Liu, Jingjing Chen, and Libo Qin. 2025.
\newblock Dlpo: Towards a robust, efficient, and generalizable prompt optimization framework from a deep-learning perspective.
\newblock \emph{arXiv preprint arXiv:2503.13413}.

\bibitem[{Perez and et~al.(2021)}]{perez2021true}
Ethan Perez and et~al. 2021.
\newblock True few-shot learning with language models.
\newblock \emph{arXiv preprint arXiv:2105.11447}.

\bibitem[{Pryzant et~al.(2023)Pryzant, Iter, Li, Lee, Zhu, and Zeng}]{pryzant-etal-2023-automatic}
Reid Pryzant, Dan Iter, Jerry Li, Yin Lee, Chenguang Zhu, and Michael Zeng. 2023.
\newblock \href {https://doi.org/10.18653/v1/2023.emnlp-main.494} {Automatic prompt optimization with {\textquotedblleft}gradient descent{\textquotedblright} and beam search}.
\newblock In \emph{Proceedings of the 2023 Conference on Empirical Methods in Natural Language Processing}, pages 7957--7968, Singapore. Association for Computational Linguistics.

\bibitem[{Srivastava and Yao(2025)}]{srivastava2025revisiting}
Saurabh Srivastava and Ziyu Yao. 2025.
\newblock Revisiting prompt optimization with large reasoning models-a case study on event extraction.
\newblock \emph{arXiv preprint arXiv:2504.07357}.

\bibitem[{Suzgun et~al.(2022)Suzgun, Scales, Sch{\"a}rli, Lewkowycz, Stenmark, Yao, Yu, Austin, Chowdhery, Le et~al.}]{suzgun2022challenging}
Mirac Suzgun, Nathan Scales, Nathanael Sch{\"a}rli, Aitor Lewkowycz, Mikael Stenmark, Shunyu Yao, Adams Yu, Jacob Austin, Aakanksha Chowdhery, Quoc Le, and 1 others. 2022.
\newblock \href {https://arxiv.org/abs/2210.09261} {Challenging big-bench tasks and whether chain-of-thought can solve them}.
\newblock \emph{arXiv preprint arXiv:2210.09261}.

\bibitem[{Team(2024)}]{qwen2024qwen25}
Qwen Team. 2024.
\newblock \href {https://arxiv.org/abs/2412.15115} {Qwen2.5 technical report}.
\newblock \emph{arXiv preprint arXiv:2412.15115}.

\bibitem[{Wang et~al.(2024)Wang, Li, Wang, Bai, Luo, Zhang, Jojic, Xing, and Hu}]{wang2024promptagent}
Xinyuan Wang, Chenxi Li, Zhen Wang, Fan Bai, Haotian Luo, Jiayou Zhang, Nebojsa Jojic, Eric Xing, and Zhiting Hu. 2024.
\newblock \href {https://openreview.net/forum?id=22pyNMuIoa} {Promptagent: Strategic planning with language models enables expert-level prompt optimization}.
\newblock In \emph{The Twelfth International Conference on Learning Representations}.

\bibitem[{Xiang et~al.(2025)Xiang, Zhang, Yu, Teng, Tu, Liang, Hong, Wu, and Luo}]{xiang2025self}
Jinyu Xiang, Jiayi Zhang, Zhaoyang Yu, Fengwei Teng, Jinhao Tu, Xinbing Liang, Sirui Hong, Chenglin Wu, and Yuyu Luo. 2025.
\newblock Self-supervised prompt optimization.
\newblock \emph{arXiv preprint arXiv:2502.06855}.

\bibitem[{Yang et~al.(2024)Yang, Wang, Lu, Liu, Le, Zhou, and Chen}]{yang2024large}
Chengrun Yang, Xuezhi Wang, Yifeng Lu, Hanxiao Liu, Quoc~V Le, Denny Zhou, and Xinyun Chen. 2024.
\newblock \href {https://openreview.net/forum?id=Bb4VGOWELI} {Large language models as optimizers}.
\newblock In \emph{The Twelfth International Conference on Learning Representations}.

\bibitem[{Ye et~al.(2023)Ye, Chen, Xu, Zu, Shao, Liu, Cui, Zhou, Gong, Shen et~al.}]{ye2023comprehensive}
Junjie Ye, Xuanting Chen, Nuo Xu, Can Zu, Zekai Shao, Shichun Liu, Yuhan Cui, Zeyang Zhou, Chao Gong, Yang Shen, and 1 others. 2023.
\newblock \href {https://arxiv.org/abs/2303.10420} {A comprehensive capability analysis of gpt-3 and gpt-3.5 series models}.
\newblock \emph{arXiv preprint arXiv:2303.10420}.

\bibitem[{Zellers et~al.(2019)Zellers, Holtzman, Bisk, Farhadi, and Choi}]{zellers2019hellaswag}
Rowan Zellers, Ari Holtzman, Yonatan Bisk, Ali Farhadi, and Yejin Choi. 2019.
\newblock \href {https://arxiv.org/abs/1905.07830} {Hellaswag: Can a machine really finish your sentence?}
\newblock In \emph{Proceedings of the 57th Annual Meeting of the Association for Computational Linguistics}, pages 4791--4800.

\bibitem[{Zhang and Sang(2025)}]{Zhang_Sang_2025}
Yuxiang Zhang and Jitao Sang. 2025.
\newblock \href {https://doi.org/10.1609/aaai.v39i24.34794} {Encoder of thoughts: Enhancing planning ability in language agents through structural embedding}.
\newblock \emph{Proceedings of the AAAI Conference on Artificial Intelligence}, 39(24):25994--26002.

\bibitem[{Zhao et~al.(2021)Zhao, Wallace, Wang, and et~al.}]{zhao2021calibrate}
Zhengxuan Zhao, Eric Wallace, Shi Wang, and et~al. 2021.
\newblock Calibrate before use: Improving few-shot performance of language models.
\newblock In \emph{ICML}.

\bibitem[{Zhou et~al.(2023)Zhou, Muresanu, Han, Paster, Pitis, Chan, and Ba}]{zhou2023large}
Yongchao Zhou, Andrei~Ioan Muresanu, Ziwen Han, Keiran Paster, Silviu Pitis, Harris Chan, and Jimmy Ba. 2023.
\newblock \href {https://openreview.net/forum?id=92gvk82DE-} {Large language models are human-level prompt engineers}.
\newblock In \emph{The Eleventh International Conference on Learning Representations}.

\end{thebibliography}

\appendix
\section{Appendix}

\subsection{Justification for Selecting Eight Principles}

We selected eight principles to balance **coverage**, **interpretability**, and **practical usability**. This decision was informed by both empirical observations and established best practices in rubric design.

Educational assessment literature recommends limiting the number of evaluation dimensions to between 6 and 8 to ensure reliability and manageability in scoring. According to Stevens and Levi (2005), "more levels typically means more time spent on assessment," and a rubric should be designed to "break down a task into components and identify the importance of these components" without overwhelming the evaluator.

In our case, we began by identifying over a dozen quality dimensions commonly used across summarization, translation, instruction-following, and reasoning evaluation settings. We then merged semantically overlapping or operationally redundant criteria—such as combining factuality and logical consistency into \textit{Correctness}, or fluency and stylistic coherence into \textit{Expression Style}.

The resulting eight principles are:
\subsection{Detailed Criteria of Eight Principles}
Table~\ref{tab:principles_detail} presents the detailed criteria of principles employed in $p_{\text{eval}}$. The subsequent guidelines elaborate on each principle, and the PCG framework generates critiques based on these criteria.

\begin{table*}[]
    \caption{Detailed Criteria of Eight Principles}
    \label{tab:principles_detail}
    \centering
    \small
    % \begin{tabular}{l|l}
    \begin{tabular}{p{3cm} | p{10cm}}
        \toprule
        principle & description \\
        \midrule
        completeness & Does the output include all key elements present in the expected output?\\
                    & Are any core ideas, steps, or facts missing compared to the expected answer? \\
        conciseness & Does the output maintain a similar level of brevity as the expected output? \\
                    & Are there unnecessary additions or repeated content beyond what is expected \\
                    & If visible reasoning is expected or allowed by the task, do not penalize the output for justified length due to reasoning steps. Only penalize verbosity that is unrelated to the task objective or that repeats content unnecessarily.\\
        correctness & Does the final output match the correct result, based strictly on factual or logical correctness?\\
                    & Do not consider the reasoning or explanation here—only whether the final output is correct and aligned with task constraints.\\
                    & For fixed-format tasks or tasks requiring structured answers, the final answer must match the expected output exactly in format, content, and position (e.g., on a separate line if required) \\
        expression style & Does the output follow the format, tone, and structure shown in the expected output? \\
                    & Are there unnecessary differences in sentence style, layout, or tone?\\
        faithfulness & Does the output avoid adding content not present in the expected output?\\
                    & Are all statements supported by the original question and context?\\
        meaning accuracy & Does the output convey the same intended meaning as the expected output? \\ 
                        & Is the reasoning process logically consistent with the way the expected output addresses the task? \\
        reasoning quality & Is the reasoning process logically valid, step-by-step, and aligned with the task intent? \\
   & Are intermediate steps necessary, accurate, and well-structured? \\
   & If the prompt expects visible reasoning, ensure it is included in the output and forms a logically coherent path to the answer.\\
        structural alignment & Does the output follow the expected structural organization (e.g., headline-body separation, bullet points, code block structure)?\\
        & Are the sections, hierarchy, or formatting explicitly aligned with the expected style?  \\
        & If the task expects visible reasoning followed by a final answer, check that the reasoning precedes the final answer and that the final answer is clearly isolated (e.g., on a separate line and in the required format). The final answer must appear in the same structure and format as shown in the expected output. \\   
        \bottomrule
    \end{tabular}
\end{table*}

\subsection{Optimized Prompt for SAMsum}
Table~\ref{tab:prompt-samsum} shows the prompt, identified by ZERA from the experiment in Section 4.2.1. The system and user prompts are adapted by including task input context. 

\begin{table*}[t]
\centering
\caption{Optimized Prompt on SAMSum Task. In this optimization, LLaMA-3.1-70B-Instruct is used for PCG and MPR. The same model is utilized as a task LLM. The reported performance in Table 3 can be easily reproducible with the following prompt.}
\label{tab:prompt-samsum}
\small
\begin{tabular}{p{3cm} | p{10cm}}
\toprule
Prompt Type & Content \\
\midrule
System Prompt & \texttt{
You are an expert in crafting structured summaries from conversational text. Your task is to distill the conversation into a single, clear sentence, highlighting crucial factual elements like who, what, where, and when. Avoid adding interpretations or including emotional content unless it is directly stated.} \\
\midrule
User Prompt & \texttt{Carefully read the given conversation. Extract the core facts into a single concise sentence summary, ensuring you include who, what, where, and when. Stick to information explicitly stated and refrain from adding personal emotions or relationships unless directly mentioned. TASK HINTS Focus on clear and directly stated facts. Do not infer or fill in gaps unless explicitly prompted by the conversation. Use a single sentence format to convey all necessary details. FEW SHOT EXAMPLES Example 1  Question Dorothy Happy anniversary to you and Sarah!! conversation continues...  Answer Damian and Sarah are celebrating their 17th anniversary in Zakopane. Example 2 Question Madelene pizza 5 o'clock? conversation continues...  Answer Madelene and John will meet for pizza and prosecco at their usual place at 5 pm.  Example 3  Question Tory guys, I need your help conversation continues...   Answer Tim will borrow 3 books for Tory. Ensure your summary is succinct and captures all critical factual details to match the example structure.}
\end{tabular}
\end{table*}

\subsection{Optimized Prompt for Epstemic Task in BBH}
Table~\ref{tab:prompt-bbh-epistemic} shows the prompt, identified by ZERA from the experiment in Section 4.2.1. The system and user prompts are adapted by including task input context. 

\begin{table*}[t]
\centering
\caption{Optimized Prompt on BBH - epistemic Task. In this optimization, LLaMA-3.1-70B-Instruct is used for PCG and MPR. The same model is utilized as a task LLM. The reported performance in Table 3 can be easily reproducible with the following prompt.}
\label{tab:prompt-bbh-epistemic}
\small
\begin{tabular}{p{3cm} | p{10cm}}
\toprule
Prompt Type & Content \\
\midrule
System Prompt & \texttt{
You are an expert at solving logical deduction puzzles related to truth-tellers and liars. Reason naturally and freely through each puzzle, exploring logical relationships step-by-step without constraints. Only after fully completing your logical analysis, clearly and succinctly state your conclusion in the exact format: Final Answer: Yes or Final Answer: No } \\
\midrule
User Prompt & \texttt{Analyze the given statements carefully and determine if the indicated individual tells the truth. Clearly reason step-by-step, explicitly stating after each deduction whether each individual \"tells the truth\" or \"lies\". Conclude clearly. Example 1: Question: Alejandro lies. Amberly says Alejandro tells the truth. Osvaldo says Amberly lies. Vernell says Osvaldo lies. Shenna says Vernell lies. Does Shenna tell the truth? Reasoning:1. Alejandro lies (given); Alejandro lies.2. Amberly claims Alejandro tells the truth; thus, Amberly lies.3. Osvaldo says Amberly lies, which is accurate; therefore, Osvaldo tells the truth.4. Vernell claims Osvaldo lies, but this is false; Vernell lies.5. Shenna correctly says Vernell lies; Shenna tells the truth.Final Answer: Yes Example 2: Question: Delbert tells the truth. Delfina says Delbert lies. Antwan says Delfina tells the truth. Helene says Antwan lies. Sima says Helene lies. Does Sima tell the truth?Reasoning:1. Delbert tells the truth (given); Delbert tells the truth.2. Delfina claims Delbert lies, making Delfina's claim false; therefore, Delfina lies.3. Antwan says Delfina tells the truth, but Delfina lies; thus, Antwan lies.4. Helene says Antwan lies, which is accurate; Helene tells the truth.5. Sima claims Helene lies, but Helene is truthful; therefore, Sima lies.Final Answer: No}
\end{tabular}
\end{table*}

\subsection{Primary Prompts of Benchmarks}

Table~\ref{tab:baseline-prompts} shows the primary prompts of Benchmarks. The baseline performance used over the main paper indicate the task performance utilizing the primary prompts.

\begin{table*}[t]
\centering
\caption{Minimal baseline prompts used for each dataset. These prompts are deliberately simple, designed only to meet standard evaluation criteria such as format compliance, without optimization or handcrafted instruction engineering.}
\small
\label{tab:baseline-prompts}
\renewcommand{\arraystretch}{1.3}
\begin{tabular}{p{3cm} | p{10cm}}
\toprule
Dataset & Baseline Prompt \\
\midrule
GSM8K & Provide the final answer prefixed with "\#\#\#\#". Do not include any explanation. \\
MMLU / MMLU-Pro & Choose the best answer from the options A–D. Answer using only the option letter in parentheses. \\
BBH & Choose the correct option from A–J. Return only the final answer enclosed in parentheses. \\
CNN/DailyMail / SAMSum & Summarize the passage below in 3–5 sentences. Be concise. \\
MBPP & Complete the function definition to pass all test cases. Output only the completed function code. \\
HumanEval & Implement the function as described. Return only executable Python code. \\
HellaSwag & Select the most plausible ending (A–D). Return only the correct letter. \\
\bottomrule
\end{tabular}
\end{table*}

\subsection{Prompt Evolution Examples of ZERA}

Table~\ref{tab:prompt-evolution-gsm8k} and~\ref{tab:prompt-evolution-samsum} show another examples of prompt evalution. They start zero initialization but improve the instruction and guidelines by observing task samples in the lens of the principles. 

\begin{table*}[h]
\centering
\caption{Prompt Structure Evolution Across Iterations (Example: GSM8K)}
\label{tab:prompt-evolution-gsm8k}
\small
\renewcommand{\arraystretch}{1.2}

\begin{tabular}{c|p{5cm}|p{6cm}}
\toprule
Iteration & System Prompt & User Prompt \\
\midrule
1 & You are a helpful assistant. & Hello! I'm here to help you. \\
2 & You are an expert problem solver who provides clear and concise reasoning before stating the final answer. & For each math problem, carefully walk through the reasoning step-by-step to solve it. At each calculation step, make sure to show your work using inline explanations with calculations in the format <<operation=result>>. Once the reasoning is complete, present the final answer on a separate line, formatted with \'\#\#\#\#\' before the number to match the expected output structure. [ZERA-generated reasoning exemplar, 2-shot] By following this guide, focus on allowing natural reasoning while ensuring the output format meets the needed structure. \\
10 & \raggedright \textbf{You are an expert math problem solver specialized in breaking down complex problems through clear and detailed step-by-step reasoning. Ensure logical coherence and mathematical precision in every explanation. Emphasize transparency and clarity in your reasoning to maintain focus on deriving correct conclusions.} & \textbf{For each math problem, walk through the solution process step-by-step, detailing each calculation and logical inference. Use inline explanations in angle brackets (e.g., `<<operation=result>>`) to clarify each operation and intermediate result. Conclude your solution with the final answer presented on a new line starting with \'\#\#\#\#\' to highlight the answer distinctly. Maintain clarity and conciseness throughout the explanation.[ZERA-generated reasoning exemplar, 1-shot] By following this guide, maintain natural reasoning while ensuring the final output aligns with the required structure. Focus on logical flow and seamless progression toward deriving the proper conclusion.} \\
\bottomrule
\end{tabular}
\vspace{0.5em}
\par\smallskip
\small\textit{Shown: GSM8K dataset. Prompt refinement progresses toward structured, evaluation-aligned formats. At later stages, ZERA introduces self-generated reasoning exemplars (e.g., 1-shot) tailored to task feedback.}
\end{table*}

\begin{table*}[h]
\centering
\caption{Prompt Structure Evolution Across Iterations (Example: SAMSum)}
\label{tab:prompt-evolution-samsum}
\small
\renewcommand{\arraystretch}{1.2}

\begin{tabular}{c|p{5cm}|p{6cm}}
\toprule
Iteration & System Prompt & User Prompt \\
\midrule
1 & You are a helpful assistant. & Hello! I'm here to help you. \\
2 & You are an AI assistant skilled at producing concise, factual summaries of conversations. Summarize accurately using only explicit details, avoiding speculation and inference about unstated motivations or beliefs. & Summarize the following conversation in a single concise paragraph, clearly stating only the explicitly mentioned facts and key details. Do not speculate about unmentioned reasons, emotions, or motivations. [ZERA-generated reasoning exemplar, 2-shot] Now summarize this conversation: \\
6 & \raggedright \textbf{You are an AI assistant adept at accurately summarizing short conversations. Focus solely on explicitly mentioned factual details such as people's names, specific items, tasks to perform, exact locations, precise time references, and explicit instructions. Strictly avoid speculation, inference, humor, or assumptions about unstated motivations or implicit meanings. Provide summaries that are concise, factual, and explicitly reflect only the provided conversation.} & \textbf{Summarize the following conversation explicitly, accurately, and concisely. Clearly state only explicitly mentioned information and include specific people, items, explicit tasks requested, exact locations, and precise instructions or timelines. Do not speculate or infer unstated emotions, motivations, or beliefs. [ZERA-generated reasoning exemplar, 3-shot] Now summarize this conversation explicitly and concisely. Explicitly identify people, clearly stated locations, explicitly requested items or tasks, and timelines.  Avoid speculation, inference, humor, or emotional interpretation not explicitly mentioned. Double-check exact locations explicitly stated to avoid confusion or misreporting.  Preserve explicit ordering of requested tasks and instructions.} \\
\bottomrule
\end{tabular}
\vspace{0.5em}
\par\smallskip
\small\textit{Shown: GSM8K dataset. Prompt refinement progresses toward structured, evaluation-aligned formats. At later stages, ZERA introduces self-generated reasoning exemplars (e.g., 1-shot) tailored to task feedback.}
\end{table*}

\end{document}